\documentclass[journal]{IEEEtai}

\usepackage[colorlinks,urlcolor=blue,linkcolor=blue,citecolor=blue]{hyperref}
\usepackage{color,array}
\usepackage{graphicx}
\usepackage{amsmath,amssymb,amsthm,bm,bbm}
\usepackage{color,xcolor,colortbl,soul}
\usepackage{graphicx,float,caption,subfig,hhline}
\usepackage{cite}
\usepackage{cleveref}
\usepackage{algorithm,algorithmic,multirow,MnSymbol}
\usepackage{setspace}
\usepackage{etoolbox}
\AtBeginEnvironment{algorithmic}{\onehalfspacing}

\allowdisplaybreaks[4]

\newtheorem{thm}{Theorem}
\newtheorem{dfn}{Definition}
\newtheorem{lem}{Lemma}
\DeclareMathOperator{\diag}{diag}

\begin{document}

	\title{Dynamic Instance--Wise Classification in Correlated Feature Spaces}

	\author{Yasitha~Warahena Liyanage, \IEEEmembership{Student member, IEEE}, Daphney--Stavroula~Zois, \IEEEmembership{Member, IEEE}, and Charalampos~Chelmis, \IEEEmembership{Member, IEEE}
		\thanks{Manuscript received (date to be filled by Editor). This material is based upon work supported by the National Science Foundation under Grants ECCS--1737443 \& CNS--1942330.}
		\thanks{The material in this paper substantially extends the conference paper \cite{liyanage2020on-the-fly} by providing novel theoretical results on the structure of the optimum solution, complete proofs and detailed experimentation on $11$ datasets across various domains.}
		\thanks{Y. Warahena Liyanage and D.--S. Zois are with the Department of Electrical and Computer Engineering, University at Albany, SUNY, NY, 12222 USA (e--mail: yliyanage@albany.edu, dzois@albany.edu).}
		\thanks{C. Chelmis is with the Department of Computer Science, University at Albany, SUNY, NY, 12222 USA (e-mail: cchelmis@albany.edu).}
		\thanks{The Associate Editor coordinating the review of this manuscript and approving it for publication was (name to be filled by Editor).}}
	
	\markboth{Journal of IEEE Transactions on Artificial Intelligence, Vol. 00, No. 0, Month 2021}
	{Y. Warahena Liyanage \MakeLowercase{\textit{et al.}}: Dynamic Instance--Wise Joint Feature Selection and Classification}
	
	\maketitle
	
	\begin{abstract}
		In a typical supervised machine learning setting, the predictions on all test instances are based on a common subset of features discovered during model training. However, using a different subset of features that is most informative for each test instance individually may not only improve prediction accuracy, but also the overall interpretability of the model. At the same time, feature selection methods for classification have been known to be the most effective when many features are irrelevant and/or uncorrelated. In fact, feature selection ignoring correlations between features can lead to poor classification performance. In this work, a Bayesian network is utilized to model feature dependencies. Using the dependency network, a new method is proposed that sequentially selects the best feature to evaluate for each test instance individually, and stops the selection process to make a prediction once it determines that no further improvement can be achieved with respect to classification accuracy. The optimum number of features to acquire and the optimum classification strategy are derived for each test instance. The theoretical properties of the optimum solution are analyzed, and a new algorithm is proposed that takes advantage of these properties to implement a robust and scalable solution for high dimensional settings. The effectiveness, generalizability, and scalability of the proposed method is illustrated on a variety of real--world datasets from diverse application domains.
	\end{abstract}
	
	\begin{IEEEImpStatement}
The ability to rationalize which features to use to classify each data instance is of paramount importance in a wide range of application domains, including but not limited to medicine, criminal justice, and cybersecurity. Correlations between features, and the need for variable selection at the same stage as classification, in such application domains present additional challenges to machine learning related to classification accuracy and computationally intractability. The proposed framework presents, to the best of our knowledge, the first practical solution that balances between classification accuracy and sparsity at the instance level, by dynamically choosing the most informative features, relative to each instance, from a set of potentially \textit{correlated} features, for classifying each \textit{individual} instance. The proposed framework achieves reductions up to $82\%$ in the average number of features used by state--of--the--art methods without sacrificing accuracy, and is robust for up to $10\%$ of missing features. Broad positive societal implications include: (1) fast, accurate and cost--efficient inference in complex dynamic settings, and (2) ease of interpretation of and trust in machine learning outcomes by domain experts (e.g., doctors, lawyers).
	\end{IEEEImpStatement}
	
	
	\section{Introduction}\label{sec:introduction}
	
	\IEEEPARstart{A} ~wide range of applications, including but not limited to medicine and robotics, demand practical solutions that can perform feature selection and classification jointly in a dynamic setting for each data instance individually. For example, consider a scenario, where a doctor is called to provide a medical diagnosis to a patient. The doctor's diagnosis may often be time--critical (e.g., in emergencies) and/or depend on numerous costly medical tests (out of which features are to be extracted), some of which cost thousands of dollars \cite{diagcost}. At the same time, a \textit{different} set of tests may be appropriate for \textit{each} individual patient (i.e., data instance). For instance, \cite{kao2015characterization} has shown that relevant features for predicting heart failure may differ across patient subgroups. Finally, considering dependencies between medical tests (and corresponding features) is equally important \cite{zhao2009searching}, since individual features may seem irrelevant with the class when examined independently, but when combined may improve classification accuracy and at the same time enhance interpretability of the final decision. Similarly, in the domain of robotics, an autonomous vehicle can control the view of its environment (e.g., change position, modify sensors parameters) to inspect and classify an object of interest \cite{hollinger2017active}. In this context, it may be important to select which sensors to use or what kind of measurements to take, while at the same time ensuring that an object in the field of view can be accurately classified.

	In this article, the problem of selecting \textit{which features} to use to \textit{classify each test instance} as features \textit{sequentially} arrive one at a time is explored. The current work extends our prior work~\cite{liyanage2020fly}, which formalized this problem with the simplifying assumption that features are conditionally independent given the class variable. Specifically, building upon~\cite{liyanage2020fly}, herein features ensure consistency: correlation vs dependency are efficiently and effectively modeled using a Bayesian network. In general, handling feature dependencies can lead to computationally intractable solutions \cite{molnar2020interpretable}. Therefore, a novel feature ordering is presented, such that each selected feature contains the maximum possible new information about the class variable with respect to the already evaluated feature set. The optimum solution is also derived for this more general formulation and important structural properties are analyzed that facilitate the design of a scalable method. The proposed method is evaluated and compared with prior work (including our own prior work~\cite{liyanage2020fly}) on a variety of real--world datasets.
	
	
	Next, the unique contributions of this work are summarized as follows: i) the optimum stopping feature (i.e., the feature at which the sequential evaluation process terminates) and the optimum classification strategy are mathematically derived for each data instance individually without imposing any assumptions on feature dependencies; ii) the structure of the optimum solution is theoretically analyzed; iii) an efficient implementation of the optimum solution is introduced that considers feature correlations to guide the feature selection process; and iv) the effectiveness, generalizability, and scalability of the proposed method are evaluated using eleven publicly available datasets. To facilitate reproducibility, the source code of the proposed method will be made available on GitHub upon acceptance of this manuscript.

	\section{Related Work}
	
	In this section, the most relevant prior work on feature selection and classification is summarized.

	To accommodate large or unknown feature spaces during \textit{model training}, streaming feature selection methods~\cite{perkins2003online, zhou2005streaming, wu2012online, yu2014towards, zhou2019ofs, hu2018survey, zhou2019online} are designed to handle features arriving sequentially over time. Existing work on this area can be roughly categorized into two directions depending on the availability of prior information about the feature space~\cite{perkins2003online, zhou2005streaming, wu2012online, yu2014towards} or not~\cite{zhou2019ofs}. In general, various threshold--based approaches have been proposed, where a newly arriving feature is selected if a constraint is satisfied (e.g., predefined threshold~\cite{perkins2003online}, dynamically varying threshold~\cite{zhou2005streaming}, conditional independence via $G^2$--test~\cite{wu2012online} or Fisher’s $Z$--test~\cite{yu2014towards}). Motivated by real--world applications, where training instances arrive sequentially or access to the full training dataset is not feasible,~\cite{wu2017large} jointly trains a linear model and acquires a sparse representation of the feature space \textit{global} to the entire dataset during the training process. All the above works use the \textit{same fixed} set of features to classify \textit{all instances} during \textit{testing}. In sharp contrast to the above lines of work, in the proposed setting, both the training instances and the full feature space is available during the training process, and the goal is to \textit{dynamically} select and use for classification a \textit{different} subset of features for \textit{each instance} during the \textit{testing process}. As a result, each testing instance is classified using \textit{different variable features}. This instance--wise property is demonstrated in Section~\ref{sec:instance-wise_feature_selection}.
	
	Recent studies have shown that relevant features may differ across data instances, for example, in predicting heart failure for different patients~\cite{kao2015characterization}. At the same time, as complex machine learning models become more prevalent, the need to interpret their results becomes critical. Hence, instance--wise feature selection~\cite{chen2018learning,yoon2018invase,xiao2019mixture} tries to identify a small number of relevant features that explain/predict the output of a machine learning model during testing. These methods work in a \textit{static} setting, since all feature values of a test instance must be first revealed. Such methods do not scale for large feature spaces, since the search space grows exponentially with the number of features. The method proposed herein can be used for model interpretability, and as such is related to these methods. However, unlike that line of work, the proposed method is \textit{dynamic}, in the sense that features arrive sequentially one at a time during testing and the goal is to \textit{jointly} select features and classify each data instance in this regime. Additionally, the number of features used for each instance is neither fixed nor predefined; it is \textit{optimally} derived by the proposed framework. Finally, the proposed method scales well with large feature spaces being able to accommodate more than 1 million features.
	

Similar to this work, classification with costly features~\cite{dulac2011datum, janisch2019classification,janisch2020classification} considers costs associated with feature evaluation and misclassification and the goal is to limit the number of features used for classification per data instance during testing. Such methods, however, define the problem \textit{globally} with respect to the training dataset, namely, by introducing a penalization term to limit the number of features used for classification in a standard empirical loss minimization problem. In that sense, even though such approaches end up sequentially evaluating different features per instance before they classify it, the resulting classification function is \textit{globally} learned with respect to the \textit{dataset}. This modified problem is shown to be equivalent to a deterministic Markov decision process (MDP) formulation and solved by a linear~\cite{dulac2011datum} or non-linear approximation~\cite{janisch2019classification,janisch2020classification} of the associated Q--function. The size of the state space of the MDP grows exponentially with the dimension of the feature space, making these methods impractical for high--dimensional settings. The proposed approach in this article is conceptually different from these methods in that the problem of joint feature selection and classification is defined and solved \textit{individually} with respect to \textit{each data instance}. In addition, the optimum classification strategy and the optimum number of features to be used for classification for each data instance individually are mathematically derived in the generic case of correlated features. Last but not least, key properties of the optimum solution are uncovered, thus enabling the proposed method to scale to large feature spaces.

	\section{Proposed Framework} \label{sec:PF}
	
	In this section, the task of instance--wise supervised multi--class classification in correlated feature spaces is posed as a sequential decision process for which the optimum solution is derived. Specifically, the goal is to learn to sequentially choose a subset of features, relative to each test instance, using which each particular instance is to be assigned to one of $L$ classes. Table \ref{tbl:symbols} summarizes the notation used hereafter.

\begin{table}[t]
	\centering
	\caption{Explanation of main symbols used in this article.}
	\vspace{-0.1in}
	\label{tbl:symbols}
	\begin{footnotesize}
		\begin{tabular}{|l|p{6cm}|}
			\hline
			Symbol     & Description  \\  \hline
			$S$ & collection of data instances\\ \hline
			$s$ & data instance $s \in S$ \\ \hline
			$F$ & Feature set \\ \hline
			$K$ & \# of features\\ \hline
			$F_k$ & $k$th feature $F_k \in F$ ($1 \leqslant k \leqslant K$)\\\hline
			$f_k$ & value of $k$th feature $F_k$ ($1 \leqslant k \leqslant K$)\\\hline
			$L$ & \# of classes\\ \hline
			$\mathcal{C}$ & class variable \\ \hline
			$c_{i}$ & class assignment to class $\mathcal{C}$ \\\hline
			$p_{i}$ & prior probability of class $i$ \\ \hline
			$e_k$ &  Cost coefficient of $k$th feature $F_k$ \\ \hline
			$Q_{ij}$ & Missclassification cost for classes $c_j$ and $c_j$ \\ \hline
		\end{tabular}
		\vspace{-0.1in}
	\end{footnotesize}
\end{table}

\begin{figure}[!tb]
		\centering
		\includegraphics[width=2.5in]{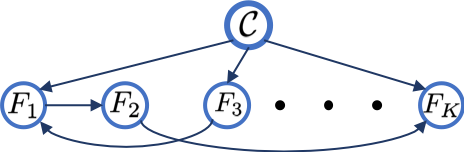}
		\caption{Sample Bayesian Network of features in $F$ and the class variable in $\mathcal{C}$.}
		\label{fig:network}
	\end{figure}

Dependencies between features and the class variable (see Fig.~\ref{fig:network}) are modeled using a Bayesian network $\mathcal{B} \triangleq \{G,\Theta\}$ \cite{pearl1988probabilistic}, which can be learned using methods such as \cite{neapolitan2004learning, koller2009probabilistic}. Specifically, a directed acyclic graph $G=(V,E)$, where $V = F \cup \{\mathcal{C}\}$ is the set of features in $F$ augmented with variable $\mathcal{C}$, and $E$ is the set of edges denoting correlations\footnote{Correlation is measured using mutual information, which quantifies the ``distance'' from independence between a pair of random variables~\cite{cover2012elements}.} among the nodes in $V$, is given, and the set $\Theta$ of conditional probability distributions for $G$ is learned during training.
	
	The goal is to leverage feature dependencies to train an instance--wise classifier that allows the number of features used for classification to vary relative to each instance, so as to optimize the trade--off between accuracy and sparsity at the individual instance level. Specifically, in order to select one out of $L$ possible classes for each instance $s$, the proposed approach evaluates features sequentially by choosing the features that are: (i) highly correlated with the class variable, and (ii) conditionally independent with the already evaluated feature set. At each step, the proposed approach considers the cost of examining the remaining features to decide between continuing the process or if enough information is available for a classification decision to be reached. Herein, two random variables $R$ and $D_R$ are defined.
	
	\begin{dfn}
		  Random variable $R \in \{0,\dots, K\}$ denotes the number of feature evaluations before the framework terminates. The event $\{R=k\}$ represents that the framework stops after evaluating $k$ features.
	\end{dfn}
	\begin{dfn}
		Random variable $ D_{R}\in \{1,\dots, L\} $ denotes the assigned value for class variable $\mathcal{C}$ based on the information accumulated up to feature $F_R$. The event $\{D_{ \{ R=k \} }=i\}$ represents assignment of class $c_i$ using features $\{f_{1}, f_{2}, \dots, f_{k}  \}$.
	\end{dfn}

%

	The optimum value $R^{*}$ and the best class assignment $D^{*}_{R^*}$ for each data instance $s \in \mathcal{S}$ are obtained by minimizing:
	\begin{equation}\label{eq:cost_function}
	J(R,D_{R}) = \mathbb{E} \Bigg \lbrace \sum_{k = 1}^{R} e_{k}  \Bigg \rbrace + \sum_{j = 1}^{L} \sum_{i = 1}^{L} Q_{ij} P(D_{R} = j, \mathcal{C} =c_{i}),
	\end{equation}
	where $e_k>0$ is the feature evaluation cost representing the time and effort required to evaluate feature $F_k$, $Q_{ij} \geqslant 0$ is the missclassification cost of assigning class $c_j$ when class $c_i$ is true, and $P(D_{R} = j, \mathcal{C} =c_{i})$ is the joint probability of assigning class $c_j$ when class $c_i$ is true. Specifically, $\mathbb{E} \lbrace \sum_{k = 1}^{R} e_{k} \rbrace$ is the expected cost accrued due to feature evaluations, and the double summation corresponds to expected cost associated with the classification rule $D_R$.
	Thus, the optimization problem is equivalent to finding $R^{*}, D^{*}_{R^*}$, such that:
	\begin{align}\label{eq:optimization_function}
		\underset{R,D_{R}}{\text{minimize}} \ J \big(R,D_{R} \big).
	\end{align}

To solve Eq.~(\ref{eq:optimization_function}), we first identify a subset of features that is highly correlated with the class variable $\mathcal{C}$, and therefore sufficient for accurately inferring its value. We achieve this by simply finding the Markov blanket \cite{pearl1988probabilistic} $M_{\mathcal{C}}$ of $\mathcal{C}$ in $G$. We use $\mathcal{B}_\mathcal{C}$ to denote the induced subgraph of $G$ with nodes $V_{\mathcal{B}_\mathcal{C}} = M_{\mathcal{C}} \cup \{ \mathcal{C} \}$. Features in $M_{\mathcal{C}}$ are then sequentially evaluated so that at each step, the feature in the subset of currently unselected features, that provides the maximum additional information about $\mathcal{C}$ with respect to the already evaluated feature set is selected. This is achieved using Eq.~(\ref{eq:update_rule}):
\begin{equation}\label{eq:update_rule}
	\pi_{k} = \frac{\diag \big( \Delta_k(F_{k}|F_{1},\dots,F_{k-1},\mathcal{C}) \big) \pi_{k-1} }{\Delta_k^T(F_{k}|F_{1},\dots,F_{k-1},\mathcal{C}) \pi_{k-1}},
\end{equation}
where $\Delta_k(F_{k}|F_{1},\dots,F_{k-1},\mathcal{C})= [P(F_{k}| F_{1},\allowbreak \dots,\allowbreak  F_{k-1},c_1),\allowbreak \dots,\allowbreak P(F_{k}| F_{1},\dots,F_{k-1},c_L)]^T$ can be computed using exact inference algorithms (e.g., \cite{koller2009probabilistic}), $\diag(A)$ denotes a diagonal matrix with diagonal elements being the elements in vector $A$, $\pi_{0} = [p_1,p_2,\dots,p_L]^T$, and
\begin{equation}\label{eq:posteriori}
	\pi_{k} \triangleq [\pi_k^1,\pi_k^2,\dots,\pi_k^L]^T
\end{equation}
is the \textit{a posteriori probability} vector with $\pi_k^i = P(c_i | F_{1}, \dotsc, F_{k})$. In lieu of $P(\mathcal{C}= c_i | F_{1} = f_{1}, \allowbreak  \dots, \allowbreak  F_{k} = f_{k})$, $P(c_i |F_{1}, \dotsc, F_{k})$ is used hereafter to improve readability.

For illustration purposes, suppose that $F_1$ is the feature most correlated with $\mathcal{C}$. Let $\widetilde{B}_{1} \subseteq \{ M_{\mathcal{C}} \cup \mathcal{C}\}$ denote the subset of $F_1$'s Markov blanket in $\mathcal{B}_\mathcal{C}$. $F_1$ is conditionally independent to all $F_k \in \{M_\mathcal{C} - \widetilde{B}_{1} \}$ given $\widetilde{B}_{1}$. Therefore, the next feature to be evaluated should come from $\{M_\mathcal{C} - \widetilde{B}_{1} \}$. It follows, that the $k$th feature to be evaluated must belong to $\{M_\mathcal{C} - \bigcup\limits_{i=1}^{k-1} \widetilde{B}_{i} 
\}$.
	
	Next, $P(D_{R} = j,\mathcal{C} =c_i )$ can be simplified as $P(D_{R} = j,c_i ) = \mathbb{E} \left \lbrace \pi_{R}^i  \mathbbm{1}_{\lbrace D_{R} = j \rbrace} \right \rbrace$ by exploiting the fact that $x_{R} = \sum_{k = 0}^{K} x_{k} \mathbbm{1}_{\lbrace R = k\rbrace}$ for any sequence of random variables $\lbrace x_{k} \rbrace$, where $\mathbbm{1}_{A} = 1$ when $A$ occurs, and $\mathbbm{1}_{A} = 0$ otherwise.
	Thus:
	\begin{equation}\label{eq:cost_function_new}
	J(R,D_{R}) = \mathbb{E} \left \lbrace \sum_{k = 1}^{R} e_{k} + \sum_{j = 1}^{L}  Q_j^T \pi_R \mathbbm{1}_{\lbrace D_{R} = j \rbrace} \right \rbrace,
	\end{equation}
	where $Q_j \triangleq [Q_{1,j},Q_{2,j},\dots,Q_{L,j}]^T$.

	At this point, the optimum classification strategy $D_{R}^{*}$ can be obtained for any given $R$ by noting that $\sum_{j = 1}^{L}  Q_j^T \pi_R \mathbbm{1}_{\lbrace D_{R} = j \rbrace} \geqslant g(\pi_{R})$, where $g(\pi_{R}) \triangleq  \min_{1 \leqslant j \leqslant L} \big [  Q_j^T \pi_R  \big ]$. Therefore, the optimum classification strategy $D_{R}^{*}$ for any given $R$ is:
		\begin{equation}\label{eq:optimal_classification_strategy}
		D_{R}^{*} = {\arg \min}_{1 \leqslant j \leqslant L} \big [ Q_j^T \pi_R  \big ],
		\end{equation}
	\noindent which assigns the given data instance to the class yielding the minimum misclassification cost. This suggests that $J(R,D_{R}) \geqslant \min_{D_{R}} J(R, D_{R})$, which in turn implies that the cost function in Eq. (\ref{eq:cost_function_new}) can be reduced to:
	\begin{equation}
	\label{Eq.reduced_cost}
	\widetilde{J}(R) = \mathbb{E} \left \{ \sum_{k = 1}^{R} e_{k} + g(\pi_{R}) \right \}.
	\end{equation}
	Note that Eq.~(\ref{Eq.reduced_cost}) can be minimized with respect to $R \in \{ 0, \dotsc, K \}$ in at most $K+1$ stages using \textit{dynamic programming}~\cite{BertsekasDPOC05}, as shown in Theorem~\ref{thm:dp_programming}.

	\begin{thm}\label{thm:dp_programming}
		For $k = K-1, \dotsc, 0$, the function $\bar{J}_{k}(\pi_{k})$  is related to $\bar{J}_{k+1}(\pi_{k+1})$ as follows:
		\begin{align}\label{eq:dp_programming}
		\bar{J}_{k}(\pi_{k}) = & \min \big[ g(\pi_{k}), \mathcal{A}_{k} (\pi_k) \big],
		\end{align}
		\noindent
		where $ \mathcal{A}_{k}(\pi_k) \triangleq e_{k+1} + \sum_{F_{k+1}} \Delta_{k+1}^T \big(F_{k+1}|F_{1},\allowbreak  \dots, \allowbreak  F_{k},\mathcal{C} \big) \pi_{k} \allowbreak  \bar{J}_{k+1} \bigg (  \frac{\diag \big(\Delta_{k+1}(F_{k+1}|F_{1},\dots,F_{k}, \mathcal{C}) \big) \pi_{k} }{\Delta_{k+1}^T(F_{k+1}|F_{1},\dots,F_{k},\mathcal{C}) \pi_{k} }\bigg )$, and $\bar{J}_{K}(\pi_{K}) = g(\pi_{K})$. The optimum feature selection strategy is $\lbrace F_{1}, F_{2}, \dotsc, F_{R^{*}} \rbrace$, where $R^{*}$ is equal to the first $k <K$ for which $ g(\pi_{k}) \leqslant \mathcal{A}_{k}(\pi_k)$, or $R^{*} = K$ if there are no more features to be evaluated. 
	\end{thm}

\begin{figure}[!tb]
		\centering
		\includegraphics[width=\columnwidth]{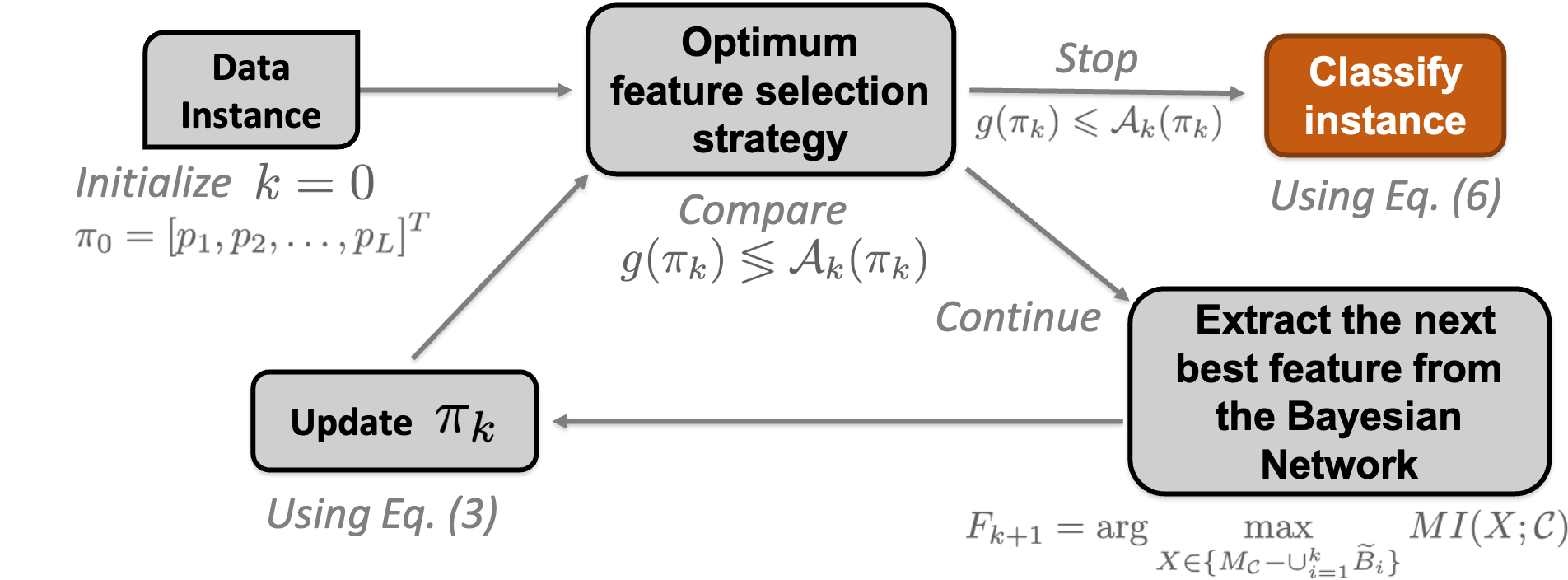} 
		\caption{Graphical illustration of the proposed framework.}
		\label{fig:framework}
	\end{figure}

Fig.~\ref{fig:framework} summarizes the process of classifying a data instance during testing. Initially, $k=0$ and $ \pi_0 = [ p_1,\dots,p_L ] $. At each stage $k$, the proposed framework compares the cost $g(\pi_k)$ of stopping to the expected cost $\mathcal{A}_k (\pi_k)$ of continuing. If $g(\pi_k) \leqslant \mathcal{A}_k (\pi_k)$, the framework stops evaluating features and classifies the instance using Eq.~(\ref{eq:optimal_classification_strategy}). Otherwise, it evaluates the next best feature $F_{k+1} = \arg \max_{X \in \{ M_{\mathcal{C}} -\cup_{i=1}^k \widetilde{B}_{i} \}} MI(X;\mathcal{C}) $ from the Bayesian network, and updates $\pi_k$ using Eq.~(\ref{eq:update_rule}). Details about the computation of mutual information, $MI(X;\mathcal{C})$, are provided in Section \ref{subsec:OS_prac}. These steps are repeated until the data instance is classified using a subset of or all $K$ features.
	
	\section{Theoretical Results}
	In this section, important properties of the optimum classification strategy $D_{R}^{*} $ in Eq.~(\ref{eq:optimal_classification_strategy}) and the optimum feature selection strategy in Eq.~(\ref{eq:dp_programming}) are analytically derived. 
	
	Consider a general form of the function $g \big(\pi_{R} \big)$ used to derive the optimum classification strategy in Eq.~(\ref{eq:optimal_classification_strategy}) given by
	$g(\varpi)\triangleq \min_{1 \leqslant j \leqslant L} \big [   Q_j^T \varpi \big ], \varpi \in [0,1]^{L},$ where $\varpi = [\omega_1,\dots,\omega_L]^T $, such that $\omega_i \geqslant 0, \sum_{i=1}^{L} \omega_i =1$. Here, the domain of $g(\varpi)$ is the probability space of $\varpi$, which is a $L-1$ dimensional unit simplex. Function $g(\varpi)$ has some interesting properties as described in Lemma~\ref{lem:g_function}. 
	
	\begin{lem}\label{lem:g_function}
		The function $g(\varpi)$ is concave, continuous, and piecewise linear. In particular, $g(\varpi)$  consists of at most $L$ hyperplanes, represented by the set $\{Q_j^T\}_{j=1}^L$ of $L$ vectors.
	\end{lem}
	
	Next, consider the general form of the function $\mathcal{A}_{k} \big(\pi_{K} \big)$ in Eq.~(\ref{eq:dp_programming}) given by 
	$\mathcal{A}_{k}(\varpi) = e_{k+1}+ \sum_{F_{k+1}}  \Delta^T \big(F_{k+1}|F_{1}, \allowbreak \dots,\allowbreak  F_{K},\mathcal{C} \big) \varpi \bar{J}_{k+1} \bigg (  \frac{\diag \big(\Delta(F_{k+1}|F_{1},\dots,F_{K},\mathcal{C}) \big) \varpi }{\Delta^T(F_{k+1}|F_{1},\dots,F_{K},\mathcal{C})\varpi }\bigg )  \bigg ].$ Lemma~\ref{lem:A_function} summarizes the key properties of this function.
	
	\begin{lem}\label{lem:A_function}
		The functions $ \mathcal{A}_{k}(\varpi), k=0,\dots,K-1$, are concave, continuous, and piecewise linear. 
	\end{lem}  
	
	The properties of functions $g(\varpi)$ and $ \mathcal{A}_{k}(\varpi)$ stated in Lemmas~\ref{lem:g_function} and~\ref{lem:A_function}, respectively, allow for a parsimonious representation of the function related to the optimum feature selection strategy in Eq.~(\ref{eq:dp_programming}) as stated in Theorem~\ref{thm:dynamic_proper}. 
	\begin{thm}\label{thm:dynamic_proper}
		At every stage $k \in \{0,\dots,K\}$, there exists a set $\{ \alpha_k^i \},\alpha_k^i \in \mathbb{R}^{1\times L}$, of vectors such that $\bar{J}_{k}(\varpi) = \min_{i}  [\alpha_k^i\varpi]$ with $\{\alpha_K^i\} \triangleq  \{Q_j^T\}_{j=0}^L$. 
	\end{thm}
	The above properties can be used to derive an efficient algorithmic implementation of the optimum solution.

	\section{Proposed Algorithm}
	Theorem~\ref{thm:dynamic_proper}, Lemma~\ref{lem:g_function} and the fact that $\bar{J}_{k}(\varpi) = \min \big[g(\varpi),\mathcal{A}_{k}(\varpi) \big],  k \in \{0,\dots,K-1\}$, allow for an efficient implementation of the optimum feature selection strategy in Theorem \ref{thm:dp_programming}. Specifically, the decision to stop or continue the feature evaluation process depends only on the vector $\alpha_k^{\mathcal{I}}  = \arg \min_{\alpha_k^i}  [\alpha_k^i\varpi]$, such that if $\alpha_k^{\mathcal{I}} \in \{Q_j^T\}_{j=0}^L$, the feature evaluation process stops, otherwise, the next feature is to be evaluated. 
	This is based on the fact that if $\arg \min_{\alpha_k^i}  [\alpha_k^i\varpi] \in \{Q_j^T\}_{j=0}^L$, it implies that $g(\varpi) \leqslant \mathcal{A}_{k}(\varpi)$ due to the following two reasons: i) the set $\{Q_j^T\}_{j=0}^L$ of $L$ vectors represents the $L$ hyperplanes of $g(\varpi)$  (see Lemma~\ref{lem:g_function}), and ii) $\bar{J}_{k}(\varpi) = \min_{i}  [\alpha_k^i\varpi] = \min \big[g(\varpi),\mathcal{A}_{k}(\varpi) \big] $. 
	Based on this fact, a dynamic \underline{I}nstance--wise joint \underline{F}eature selection and \underline{C}lassification algorithm for \underline{C}orrelated \underline{F}eatures (IFC$^2$F), is presented. Initially, $\varpi$ is set to $\pi_{0}$, and $\alpha_0^{\mathcal{I}}  = \arg \min_{\alpha_0^i}  [\alpha_0^i\varpi]$ is computed. If $\alpha_0^{\mathcal{I}} \in \{Q_j^T\}_{j=0}^L$, IFC$^2$F classifies the instance under examination to the appropriate class, based on the optimum classification strategy in Eq.~(\ref{eq:optimal_classification_strategy}). Otherwise, the first feature is evaluated. IFC$^2$F repeats these steps until either it decides to classify the instance using $<K$ features, or using all $K$ features. Algorithm~\ref{alg:example} describes these steps in detail. The input vector sets $\{\alpha_k^{i}\}$ can be computed using a standard point--based value iteration algorithm~\cite{kaelbling1998planning} during training. For simplicity, the Perseus algorithm~\cite{spaan2005perseus} is used, among
	the several point--based value iteration algorithms in the literature~\cite{shani2013survey}. Specifically, a fixed number $\beta$ (e.g., $\sim$100~\cite{spaan2005perseus}) of reachable $\varpi$ vectors from each stage, marginal probability tables $\Delta \big( F_{k}|F_{1},\dots,F_{k-1},\mathcal{C} \big)$, misclassification costs $Q_{ij}$ and feature evaluation costs $e_k$ were provided to the Perseus algorithm to obtain $\{\alpha_k^{i}\}, k \in \{0,\dots,K-1\}$.

	\begin{algorithm}[tb]
		\caption{IFC$^2$F}
		\label{alg:example}
		\begin{algorithmic}
			\STATE {\bfseries Input:} Vector sets  $\{\alpha_0^{i}\},\dots,\{\alpha_{K-1}^{i}\}$, and misclassification costs $Q_{ij},i,j \in \{1,\dots,L\}$
			\STATE {\bfseries Output:} Classification decision $D$ of the instance under examination, number $R$ of features used
			\STATE Initialize $\varpi = \pi_0$
			\FOR{$k=0$ {\bfseries to} $K$}
			\IF{$k = K$ \textbf{or} $\arg \min_{\alpha_k^i}  [\alpha_k^i\varpi] \in \{Q_j^T\}_{j=0}^L$}
			\STATE \textbf{Break}
			\ELSE 
			\STATE Obtain next feature value $f_{k+1}$
			\STATE Update $\varpi$ using Eq.~(\ref{eq:update_rule})
			\ENDIF
			\ENDFOR
			\STATE{\bfseries Return:} $D = {\arg \min}_{1 \leqslant j \leqslant L} \big [ Q_j^T \varpi  \big ]$, $R=k$		
		\end{algorithmic}
	\end{algorithm}
	
	\subsection{Practical Considerations} \label{subsec:OS_prac}
		
	The adjusted mutual information (AMI)\footnote{The maximum of this value represents perfect correlation between the variables, while a value around zero represents independence~\cite{scikit-learn}.} between each feature and the class variable is computed, and $M_\mathcal{C}$ is acquired by removing low correlation features based on a threshold $\eta$ on AMI. Specifically, $\eta$ is initialized to 1, and is iteratively halved until the number of filtered features is greater than zero. For simplicity, the Bayesian Network is assumed to exhibit a tree structure rooted at the class variable (experiments are performed on three alternative network structures in Section~\ref{sec:network_struct}). Such networks can be efficiently constructed  (e.g., by building the maximum spanning tree~\cite{friedman1996building}) using pairwise conditional mutual information. Conditional probability tables $\big($i.e., $P(F_k|\Pi_{F_k})$, where $\Pi_{F_k}$ denotes the set of parents of $F_k\big)$, are estimated using a smoothed maximum likelihood estimator. Specifically, $\hat{P}(F_a=f_a|F_b=f_b,\mathcal{C}=c_i) = \frac{N_{a,b,i} + 1}{N_{b,i} + V}$, where $N_{a,b,i}$ denotes the number of samples that satisfy $F_a =f_a$ and $F_b =f_b$, and belong to class $c_i$, $N_{b,i}$ denotes the number of samples that satisfy $F_b =f_b$, and belong to class $c_i$, and $V$ is the number of quantization levels considered. The \emph{a priori} probabilities are estimated as $P(c_i)= \frac{N_i}{\sum_{i=1}^{L}N_i},  i = 1, \dotsc, L$ where $N_i$ is the number of instances that belong to class $c_i$.  To reduce both memory requirements and the number of computations when storing and computing marginal probability tables $\Delta \big( F_{k}|F_{1},\dots,F_{k-1},\mathcal{C} \big)$, the dependency of each variable is limited to be second--order, such that the only dependency for $F_{k}$ other than $\mathcal{C}$, is its first ancestor in the set $\{F_{1},\dots,F_{k-1}\}$, except for $F_{1}$ which has $\mathcal{C}$ as its only dependent. 
	
\subsection{Complexity Analysis}\label{sec:complexity_analysis}

\noindent \textbf{Preprocessing stage:}  This stage consists of three steps. First, extracting the highly correlated feature set based on a threshold on mutual information is $\mathcal{O}(K\log K)$. Second, learning a tree--based Bayesian network with corresponding conditional probability distributions (CPDs) is $\mathcal{O}(K^2+KL)$~\cite{friedman1996building}.
Third, computing marginal probability tables from CPDs is $\mathcal{O}(KL)$. Thus, the complexity of the preprocessing stage is $\mathcal{O}(K^2+KL)$. \\
\textbf{Training stage:} In this stage, the optimum $\{\alpha_k^{i}\}$ vectors are determined. Computing the optimum $\alpha_k^i$ vectors for all $K$ stages by considering a fixed set of belief points from each stage using Perseus algorithm is $\mathcal{O}(KLV)$~\cite{spaan2005perseus}. Thus, model training is $\mathcal{O}(KLV)$. \\
\textbf{Testing stage:} The computational complexity of  computing the minimum  $\alpha_k^i\varpi$ among the set $\{\alpha_k^i\}$ is $\mathcal{O}(L)$, since: i) the  set $\{\alpha_k^{i}\}$ computed using the Perseus algorithm contains at most a constant number of vectors~\cite{spaan2005perseus}, and ii) the dot product between a pair of $[0,1]^L$ vectors require $2L-1$ computations. 
The complexity of obtaining a new feature is $\mathcal{O}(1)$, while  updating $\varpi$ using Eq.~(\ref{eq:update_rule}) is  $\mathcal{O}(L)$, since a dot product between a pair of $L$--dimensional vectors must be computed. 
Hence, IFC$^2$F can classify an instance in $\mathcal{O}(KL)$.
	
\begin{table}[t]
	\centering
	\caption{Datasets used in the experiments.}
	\vspace{-0.1in}
	\label{tbl:dataset}
	\begin{footnotesize}
		\begin{tabular}{|l|c|c|c|}
			\hline
			Dataset     & \# Instances  & \# Features &   \# Classes                \\   \hline
			Madelon     & $2,000$      &$500 $        &   $2$        \\   \hline
			Lung Cancer        &   $181$        &$12,533$      &   $2$        \\  \hline
			MLL         &   $72$         &$5,848$       &   $3$        \\  \hline
			Dexter      &   $300$        &$20,000$      &   $2$        \\  \hline
			Car         &   $174$        &$9,182$       &   $11$        \\  \hline
			Lung2       & $203$      &$3,312$       &   $5$        \\  \hline
			Leukemia    &   $72$     &$7,129$       &   $2$        \\  \hline
			Prostate    &   $102$    &$6,033$       &   $2$        \\  \hline
			Spambase    &   $4601$   &$57$          &   $2$        \\  \hline
			Dorothea    &   $800$    &$100,000$     &   $2$        \\  \hline
			News20      &   $19,996$ &$1,355,191$   &   $2$        \\  \hline
		\end{tabular}
		\vspace{-0.1in}
	\end{footnotesize}
\end{table}	

\begin{figure*}[t]
	\centering
	\includegraphics[width=0.9\textwidth]{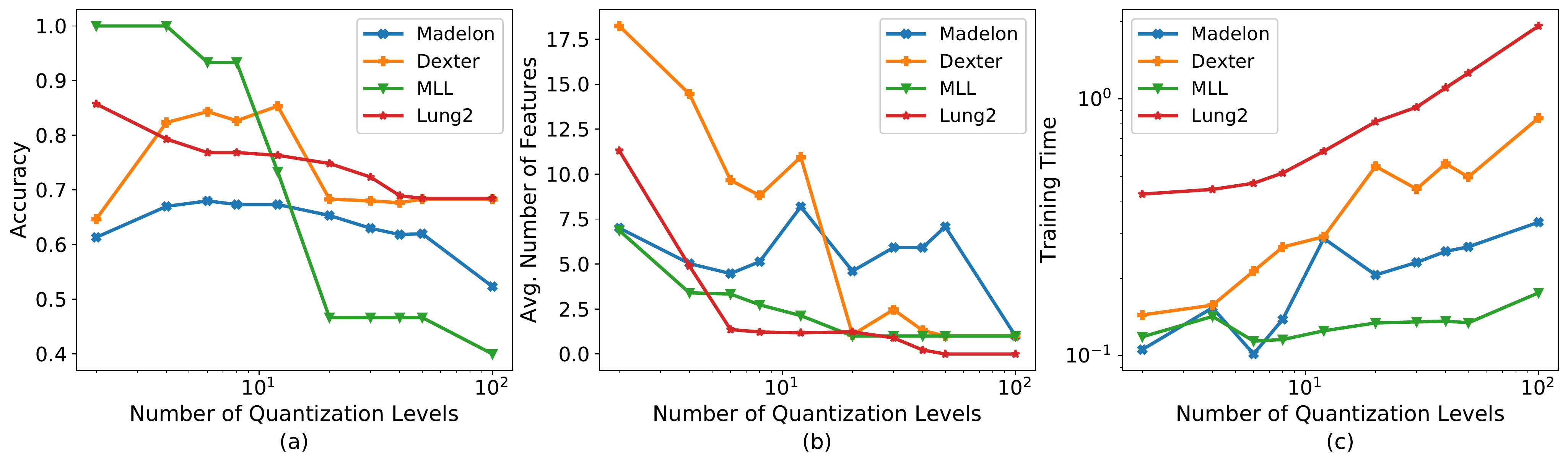} 
	\vspace{-0.1in}
	\caption{Variation of (a) accuracy, (b) average number of features, and (c) training time (sec) as a function of the number $V \in \{2, 4, 6, 8, 12, 20, 30, 40, 50, 100\}$ of bins using Lung2, Dexter, Madelon and MLL datasets.}
	\label{fig:all_bin}
	\vspace{-0.1in}
\end{figure*}
\section{Experimental Evaluation}\label{sec:Exp}

In this section, an extensive set of experiments is conducted to evaluate the performance of IFC$^2$F using $11$ benchmark datasets: 6 DNA Microarray Datasets (Lung Cancer, Lung2, MLL, Car, Leukemia, Prostate)  \cite{yang2006stable}, 4 NIPS feature selection challenge datasets (Dexter, Madelon, Dorothea, Spambase) \cite{NIPSdataset}, and 1 high dimensional dataset (News20)~\cite{Libsvm}. Table~\ref{tbl:dataset} summarizes these datasets. For Madelon, MLL, Dexter and Dorothea, the originally provided training and validation sets are used, while for the remaining datasets, five--fold cross validated results are reported. All experiments are conducted on an iMac with Quad--Core Intel Core i7 @3.30 GHz CPU, 16 GB memory, and macOS Catalina. 


\subsection{Effect of Feature Space Quantization}\label{sec:er_bin}

In Section~\ref{subsec:OS_prac}, the feature space was quantized to estimate the conditional probability tables.
Herein, the effect of the number $V$ of quantization levels on IFC$^2$F is analyzed using four datasets (Lung2, Dexter, Madelon, MLL)~(see Fig.~\ref{fig:all_bin}). It is observed that increasing the number of bins results in a significant drop in accuracy for the MLL dataset, while for the Dexter dataset, increasing the number of bins from 2 to 20 results in an improvement in accuracy. All the datasets show a reduction in the number of features used as $V$ is increased.  These observations suggest that increasing the resolution of the feature space to a very high value can cause overfitting. However, at the same time, it can help to  accommodate data sparsity in sparse datasets~(e.g., Dexter). On the other hand,
the linear relationship between training time and the number
of bins $V$ in Fig.~\ref{fig:all_bin}(c) validates the $\mathcal{O}(KLV)$ complexity of IFC$^2$F's training stage (see Section~\ref{sec:complexity_analysis}).
 In the rest of the experiments, to eliminate overfitting, $V$ is set to a moderate value (i.e., 4), except for sparse datasets (i.e., Dexter, Dorothea, Spambase and News20), where $V$ is set to a slightly higher value (i.e., 10).

\begin{table*}[!t]
	\centering
	\caption{Accuracy and average number of features used by IFC$^2$F for different feature evaluation cost values using Lung2, Dexter, Madelon and MLL datasets.}
	\vspace{-0.1in}
	\label{tbl:feat_cost}
	\begin{center}
		\begin{footnotesize}
			\begin{tabular}{|c|c|c|c|c|c|c|c|c|}
				\hline							
				\multirow{2}{*}{Dataset}
				&\multicolumn{2}{c|}{$e=0.1$}   &\multicolumn{2}{c|}{$e=0.01$} &\multicolumn{2}{c|}{$e=0.001$} &\multicolumn{2}{c|}{$e=0.0001$} \\
				\cline{2-9}
				
				 & Accuracy & Avg. \# Features  & Accuracy & Avg. \# Features & Accuracy & Avg. \# Features & Accuracy &  Avg. \#  Features \\  
				\hline
				
				Madelon &$0.6217$  &$1.00$  &$0.6700$ &$5.03$     
				&$0.6867$  &$21.54$  &$0.6767$ &$42.34$     \\ \hline
				
				Dexter &$0.6467$  &$1.00$  &$0.8533$ &$10.95$     
				&$0.8233$  &$25.52$  &$0.8133$ &$31.02$     \\ \hline
				
				MLL  &$0.9333$  &$2.20$  &$1.00$ &$3.40$     
				&$1.00$  &$3.73$  &$1.00$ &$3.73$     \\ \hline
				
				Lung2 &$0.6846$  &$0.00$  &$0.8573$ &$11.30$     
				&$0.8918$  &$18.01$  &$0.8818$ &$22.30$     \\ 
				
				\hline
			\end{tabular}
		\end{footnotesize}
	\end{center}
\end{table*}

\begin{table*}[!t]
	\centering
	\caption{Accuracy and average number of features used by IFC$^2$F for different dependency structures using Lung2, Dexter, Madelon and MLL datasets.}
	\vspace{-0.1in}
	\label{tbl:feat_graph}
	\begin{center}
		\begin{footnotesize}
			\begin{tabular}{|c|c|c|c|c|c|c|c|c|c|c|c|c|c|c|c|c|}
				\hline							
				\multirow{2}{*}{Dataset} 
				&\multicolumn{2}{c|}{Tree}   &\multicolumn{2}{c|}{Line} &\multicolumn{2}{c|}{Random} &\multicolumn{2}{c|}{Clique} \\
				\cline{2-9}
				
				 & Accuracy & Avg. \# Features & Accuracy & Avg. \# Features & Accuracy & Avg. \# Features  & Accuracy & Avg. \# Features \\  
				\hline
				
				Madelon & $0.6700$ & $5.03$   
				& $0.6150$ & $4.73$ 
				& $0.6200$ & $1.54$  
				& $0.6217$ & $1.00$ \\ \hline
				
				Dexter & $0.8533$   & $10.95$
				& $0.7833$ & $6.20$ 
				& $0.7833$ & $12.02$  
				& $0.6467$ & $1.00$ \\ \hline
				
				MLL  & $1.00$ & $3.40$   
				& $1.00$ & $4.80$ 
				& $0.9333$ & $2.73$  
				& $0.4667$ & $1.00$ \\ \hline
				
				Lung2 & $0.8573$ & $11.30$   
				& $0.8672$ & $12.57$ 
				& $0.9116$ & $15.47$  
				& $0.7733$ & $1.00$ \\  \hline
				
			\end{tabular}
		\end{footnotesize}
	\end{center}
\end{table*}

\begin{table}[h!]
	\centering
	\caption{Computational complexity of baselines. Parameters $K$ and $N$ denote number of features and instances, accordingly.}
	\vspace{-0.1in}
	\label{tbl:complexity}
	\begin{footnotesize}
		\begin{tabular}{|l|l|}
			\hline
			Method & Computational Complexity \\  \hline
			OFS--Density~\cite{zhou2019ofs}  & $\mathcal{O}(K^2N^2 \log N)$ \\ \hline
			OFS--A3M~\cite{zhou2019online} & $\mathcal{O}(K^2N^2 \log N)$ \\ \hline
			SAOLA~\cite{yu2014towards} & $\mathcal{O}(K^2)$ \\ \hline
			OSFS~\cite{wu2012online} & $\mathcal{O}(K^2KK^K)$ \\ \hline
			Fast--OSFS\cite{wu2012online} & $\mathcal{O}(KK^K)$ \\ \hline
			Alpha--Investing~\cite{zhou2005streaming} & $\mathcal{O}(K^3)$ \\ \hline
		\end{tabular}
		\vspace{-0.1in}
	\end{footnotesize}
\end{table}

\subsection{Effect of Feature Evaluation Cost}\label{sec:feat_acc}

To study the behavior of IFC$^2$F for varying values of feature evaluation cost $e = \{0.1, 0.01, 0.001, 0.0001\}$, when all  features incur the same cost (i.e., $e_k =e, \forall k$), the accuracy and the average number of features used for classification for constant misclassification costs (i.e., $Q_{ij} =1 \forall i \neq j, Q_{ii} = 0, i,j\in \{1,\dots,L\}$) are measured~(see Table~\ref{tbl:feat_cost}). Different $e$ values result in different number of features used and levels of accuracy. Intuitively, using a small portion of the total feature set leads to low accuracy, whereas when the average number of features used increases, the performance improves dramatically. From here onwards, unless specified, results are reported for $e = 0.01$, since according to Table~\ref{tbl:feat_cost}, IFC$^2$F achieves the best tradeoff between accuracy and the average number of features used for this value.

\subsection{Effect of Bayesian Network Structure}\label{sec:network_struct}

In this subsection, the behavior of IFC$^2$F is analyzed for different dependency structures by considering three alternatives in addition to the tree--based structure (``Tree'') introduced in Section~\ref{subsec:OS_prac}. Initially, features are reverse sorted (highest value first) with respect to mutual information with the class variable and the following dependency structures are considered: i) ``Line'': a directed line graph having edges pointing outward starting from the first feature in the ordering; ii) ``Random'': a random directed acyclic graph; 
and iii) ``Clique'': a complete directed graph. Note that for all of these dependency structures, the class variable is considered as a common parent connected to all feature nodes. Table~\ref{tbl:feat_graph} provides a comparison among these four dependency structures. It is observed that for all the datasets, the ``Tree'' structure outperforms the rest by achieving the best trade-off between accuracy and the number of features used. Hence, the ``Tree'' structure is considered in the rest of the experimental analysis.

\begin{table*}[t]
	\centering
	\caption{Comparison of accuracy. The highest accuracy, and the second highest accuracy are bolded and gray--shaded, and gray--shaded, respectively. Cells are marked with `- -' if the corresponding method was unable to generate results within a cutoff time of 12 days.}
	\label{tbl:acc}
	\begin{footnotesize}
		\begin{tabular}{|l|c|c|c|c|c|c|c|c|c|}
			\hline
			Dataset &IFC$^2$F &ETANA &F--ETANA   & OFS--Density & OFS--A3M & SAOLA & Fast--OSFS & OSFS & Alpha--Investing \\
			\hline

			Madelon  & \cellcolor{gray!20}  $\textbf{0.6700}$   &\cellcolor{gray!20}  $0.6217$ &$0.5180$     &$0.5117$    &$0.5117$      &$0.5817$ &$0.5417$ &$0.5817$        & $0.6050$   \\   \hline

			Lung Cancer  & $ 0.9724$ &\cellcolor{gray!20}  $\textbf{0.9890}$  &\cellcolor{gray!20}  $\textbf{0.9890}$     &\cellcolor{gray!20}  $0.9835$    &$0.9779$      &\cellcolor{gray!20}  $\textbf{0.9890}$ &\cellcolor{gray!20}  $\textbf{0.9890}$ &$0.9724$        &$0.9613$    \\   \hline

			MLL  & \cellcolor{gray!20}  $\textbf{1.00}$  &\cellcolor{gray!20}  $\textbf{1.00}$  & $0.9467$   &\cellcolor{gray!20} $0.9600$  &$0.9067$      &$0.8667$ &$0.8000$ &$0.8000$        & $0.9333$    \\   \hline

			Dexter & \cellcolor{gray!20}  $\textbf{0.8533}$ & $0.8133$  &$0.7967$     &\cellcolor{gray!20}  $0.8527$    &$0.7375$      &$0.7800$  &$0.7800$  & $0.7967$        &$0.5000$    \\   \hline

			Car  & $0.7471$ &\cellcolor{gray!20} $0.8097$ &\cellcolor{gray!20}  $\textbf{0.8274}$    &$0.5973$    &$0.7929$      &$0.7982$ &$0.6082$ &$0.5575$        &$0.6429$    \\   \hline

			Lung2 & $0.8573$  &$0.8820$  &\cellcolor{gray!20} $0.8918$     &\cellcolor{gray!20}  $\textbf{0.9117}$    &$0.8717$    &$0.8817$ &$0.8420$ &$0.8471$        &$0.8820$    \\   \hline
			
			Leukemia & \cellcolor{gray!20} $\textbf{0.9571}$  & \cellcolor{gray!20} $\textbf{0.9571}$  & \cellcolor{gray!20} $\textbf{0.9571}$     &\cellcolor{gray!20} $0.9438$    &  $0.7914$      &$0.9295$ &$0.9295$  &$0.8867$         &$0.8324$    \\   \hline
			
			Prostate  & $0.9005$ &\cellcolor{gray!20} $\textbf{0.9310}$  &$0.9010$     &\cellcolor{gray!20} $0.9210$    & $0.8148$      &$0.8910$ &$0.8633$ &$0.8833$         &$0.9014$    \\   \hline
			
			Spambase & $0.8104$ & \cellcolor{gray!20} $0.8467$  &$0.5109$     &$0.7870$    & \cellcolor{gray!20}  $\textbf{0.8598}$      &$0.8241$ &$0.8011$ &$0.8011$        &$0.8074$    \\   \hline
			
			Dorothea & \cellcolor{gray!20} $\textbf{0.9429}$ &\cellcolor{gray!20} $0.9400$  &$0.7714$     &$0.9314$    &  $0.9314$      &$0.9114$  & \cellcolor{gray!20} $\textbf{0.9429}$ &$0.9000$       &$0.6457$    \\   \hline
			
			News20 &\cellcolor{gray!20} $0.7503$ & $0.7352$  &$0.6346$     &$--$    &$--$      &\cellcolor{gray!20}  $\textbf{0.7846}$ &$--$ &$--$        &$--$    \\   \hline
			
			\textbf{Avg. rank} & $3.41$ & $2.32$  &$4.36$     &$4.82$    &$6.27$      & $4.55$ &$6.18$ &$6.95$        &$6.14$    \\   \hline
		\end{tabular}
	\end{footnotesize}
	\vspace{-0.1in}
\end{table*}

\begin{table*}[t]
	\centering
	\caption{Comparison of average number of features used. The minimum and the second minimum average number of features used are bolded and gray--shaded, and gray--shaded accordingly. Cells are marked with `- -' if the corresponding method was unable to generate results within a cutoff time of 12 days.}
	\label{tbl:feat}
	\begin{footnotesize}
		\begin{tabular}{|l|c|c|c|c|c|c|c|c|c|}
			\hline
			Dataset &IFC$^2$F  &ETANA &F--ETANA   & OFS--Density & OFS--A3M & SAOLA & Fast--OSFS & OSFS & Alpha--Investing \\ \hline

			Madelon  & $5.03$ & $4.09$  &$55.48$   &\cellcolor{gray!20}  $\textbf{2.00}$    & \cellcolor{gray!20} $\textbf{2.00}$      &\cellcolor{gray!20}  $3.00$ & \cellcolor{gray!20} $3.00$ & \cellcolor{gray!20} $3.00$        & $4.00$    \\   \hline

			Lung Cancer &\cellcolor{gray!20}  $\textbf{1.35}$ &\cellcolor{gray!20}  $2.03$ & $6.56$     &$37.20$    &$8.40$      & $52.00$ & $6.80$ & $4.00$        &$4.60$    \\   \hline

			MLL    &\cellcolor{gray!20}  $3.40$ & $5.07$  & $14.69$     & $11.00$    &$12.00$      &$28.00$ & $5.00$ &\cellcolor{gray!20}  $\textbf{3.00}$        &$7.00$    \\   \hline

			Dexter   &$10.95$  &$12.80$ &$243.4$      & $10.00$    &$104.0$      &$21.00$ &$9.00$ &\cellcolor{gray!20}  $6.00$        &\cellcolor{gray!20}  $\textbf{1.00}$    \\   \hline

			Car  & $24.20$ & $12.90$   &  $340.20$   &\cellcolor{gray!20}  $6.80$    &$36.00$      &$41.40$ &$8.40$ &\cellcolor{gray!20}  $\textbf{5.20}$        &$24.40$    \\   \hline

			Lung2   & $11.30$  &$15.59$ &$27.91$      & $16.20$    & $18.00$     &$28.20$ &\cellcolor{gray!20}  $9.40$ &\cellcolor{gray!20}  $\textbf{5.80}$        &$34.40$    \\   \hline

			Leukemia & \cellcolor{gray!20} $\textbf{1.90}$ &\cellcolor{gray!20}  $2.08$  &$9.53$     &$4.40$    &  $13.40$      &$21.60$ &$4.60$ & $2.20$        &$3.20$    \\   \hline
			
			Prostate & $4.08$ & \cellcolor{gray!20}  $3.34$  &$10.39$     &$5.80$    & $40.20$      &$14.00$ &$3.80$ &\cellcolor{gray!20}  $\textbf{1.60}$        &$7.00$    \\   \hline
			
			Spambase & \cellcolor{gray!20} $\textbf{4.72}$ & \cellcolor{gray!20}  $7.47$  &$56.00$     &$7.60$    &  $42.20$      &$24.60$ &$33.80$ &$33.80$        &$42.60$    \\   \hline
			
			Dorothea &\cellcolor{gray!20}  $\textbf{2.29}$ & \cellcolor{gray!20}  $2.89$  &$8.10$     &$17.40$    &  $34.00$      &$32.00$ &$24.00$ &  $3.00$        &$113.0$    \\   \hline
			
			News20  & \cellcolor{gray!20}  $\textbf{43.68}$ &\cellcolor{gray!20}  $81.70$ &$4000.6$     &$--$    &$--$ & $241.8$ &$--$ &$--$        &$--$    \\   \hline
			
			\textbf{Avg. rank} & $2.91$ & $3.36$  &$7.09$     &$4.68$    &$6.86$      & $6.91$ &$4.41$ &$2.86$        &$5.91$    \\   \hline
		\end{tabular}
	\end{footnotesize}
	\vspace{-0.1in}
\end{table*}

\begin{table*}[t]
	\centering
	\caption{Comparison of time (in seconds) required for feature selection (F), classification (C), joint feature selection and classification (F+C), model training (T) and preprocessing (P). The minimum and the second minimum F+C times are bolded and gray--shaded, and gray--shaded accordingly. Cells are marked with `- -' if the corresponding method was unable to generate results within a cutoff time of 12 days.}
	\label{tbl:time}
	\begin{footnotesize}
		\begin{tabular}{|l|c|c|c|c|c|c|c|c|c|c|c|}
			\hline
			Dataset & \rotatebox[origin=c]{90}{Time } &IFC$^2$F &ETANA &F--ETANA  & \rotatebox[origin=c]{90}{Time }  & OFS--Density & OFS--A3M & SAOLA & Fast--OSFS & OSFS & Alpha--Investing \\
			
			\hline	
			\multirow{4}{*}{Madelon}
			&\multirow{2}{*}{\parbox{0.2cm}{\centering{F+ C}}}  &  &  &   &F    &$129.8$    &$151.1$      & \cellcolor{gray!20}  $0.048$ &$0.076$ &$0.080$        & \cellcolor{gray!20}  $\textbf{0.033}$    \\    
			\cline{6-12}

			& & $\multirow{-2}{*}{0.070}$ &   $\multirow{-2}{*}{0.097}$  &$\multirow{-2}{*}{0.428}$    &C    &$0.011$    &$0.010$      &\cellcolor{gray!20}  $0.010$ &$0.011$ &$0.011$        &\cellcolor{gray!20}  $\textbf{0.010}$    \\   
			\cline{2-12}

			&  T   & $0.155$  & $0.247$  &$0.748$   &\multirow{2}{*}{T}     &\multirow{2}{*}{$0.074$}    &\multirow{2}{*}{$0.075$}      &\multirow{2}{*}{$0.075$}    &\multirow{2}{*}{$0.073$} &\multirow{2}{*}{$0.073$}    & \multirow{2}{*}{$0.076$ }  \\   
			
			\cline{2-5}
			&  P  &  $ 35.58$ & $0.142$ & $0.131$ &   &    &     & & &       & \\   \hline

			\multirow{4}{*}{Lung Cancer}
			
			&\multirow{2}{*}{\parbox{0.2cm}{\centering{F+ C}}} & \cellcolor{gray!20} &\cellcolor{gray!20}    &\cellcolor{gray!20}   &F     &$12.80$    &$40.43$      & $2.465$ &  $1.279$ &$18.23$        & $0.459$    \\   
			\cline{6-12} 
			
			&  & \cellcolor{gray!20} $\multirow{-2}{*}{\textbf{0.001}} $  & \cellcolor{gray!20} \multirow{-2}{*}{$0.003$}  & \cellcolor{gray!20}   \multirow{-2}{*}{$0.003$}   &C     &$0.011$    &$0.012$      & $0.013$ & $0.010$ &$0.010$        & $0.010$    \\ 
			\cline{2-12}
			
			&  T & $0.213$&$11.79$   & $1.340$   &\multirow{2}{*}{T}    &\multirow{2}{*}{$0.071$}    &\multirow{2}{*}{$0.089$}     & \multirow{2}{*}{$0.068$}      &\multirow{2}{*}{$0.068$} &\multirow{2}{*}{$0.070$}    &\multirow{2}{*}{$0.072$}    \\   
			
			\cline{2-5}
			&  P  & $146.0$ & $1.806$ & $1.796$ &   &    &     & & &       & \\   \hline
			
			\multirow{4}{*}{MLL}
			&\multirow{2}{*}{\parbox{0.2cm}{\centering{F+ C}}} & \cellcolor{gray!20}    & \cellcolor{gray!20}  &  \cellcolor{gray!20}  &F   & $1.468$   &$5.543$      &$1.513$ &$0.564$ &$4.679$        & $0.154$    \\  
			\cline{6-12} 
			
			&  &\cellcolor{gray!20}   $\multirow{-2}{*}{\textbf{0.001}}$  & \cellcolor{gray!20}   $\multirow{-2}{*}{0.003}$   &  \cellcolor{gray!20}   $\multirow{-2}{*}{0.003}$     &C    & $0.004$    &$0.011$      &$0.013$ &$0.010$ &$0.010$        & $0.010$    \\  
			\cline{2-12}
			
			& T & $0.448$  & $24.63$   & $3.193$     &\multirow{2}{*}{T}     & \multirow{2}{*}{$0.008$}     &\multirow{2}{*}{$0.071$}      &\multirow{2}{*}{$0.069$}  &\multirow{2}{*}{$0.071$} &\multirow{2}{*}{$0.071$}  &\multirow{2}{*}{$0.073$}    \\   
			
			\cline{2-5}
			&  P  & $291.2$& $0.133$ & $0.133$ &   &    &     & & &       & \\   \hline

			\multirow{4}{*}{Dexter}
			&\multirow{2}{*}{\parbox{0.2cm}{\centering{F+ C}}}  &  \cellcolor{gray!20}  &\cellcolor{gray!20} &  &F    & $77.88$    &$48453$      &$0.747$ &$1.087$ & $2.509$        &$12.98$    \\   
			\cline{6-12} 
			
			& &\cellcolor{gray!20} \multirow{-2}{*}{$\textbf{0.054}$}   &\cellcolor{gray!20} \multirow{-2}{*}{$0.125$}  & \multirow{-2}{*}{$0.681$}  & C     & $0.048$    &$0.180$      &$0.060$ &$0.038$ & $0.033$  &$0.024$    \\   
			\cline{2-12} 
			
			& T & $ 0.290$ &$22.18$  &$2.106$   &\multirow{2}{*}{T}     & \multirow{2}{*}{$0.090$}    &\multirow{2}{*}{$0.073$}      &\multirow{2}{*}{$0.069$}  &\multirow{2}{*}{$0.067$} & \multirow{2}{*}{$0.063$}      &\multirow{2}{*}{$0.089$}    \\   
			
			\cline{2-5}
			&  P  & $35.91$ & $0.133$ & $0.135$ &   &    &     & & &       & \\   \hline

			\multirow{4}{*}{Car}
			&\multirow{2}{*}{\parbox{0.2cm}{\centering{F+ C}}}  & \cellcolor{gray!20}	 & \cellcolor{gray!20} & &F    &$8.236$    &$40.55$      &$1.155$ &$0.999$ &$13.40$        &$0.710$    \\   
			\cline{6-12} 
			& & \cellcolor{gray!20}	 \multirow{-2}{*}{$\textbf{0.029}$}  &     
			
			\cellcolor{gray!20}	\multirow{-2}{*}{$0.044$} & \multirow{-2}{*}{$0.395$} &C     &$0.011$    &$0.013$      &$0.013$ &$0.010$ &$0.010$        &$0.013$    \\    
			\cline{2-12}		
			
			&T  & $4.950$ & $3059.5$ &  $37.21$  &\multirow{2}{*}{T}       &\multirow{2}{*}{$0.070$}    &\multirow{2}{*}{$0.066$}      &\multirow{2}{*}{$0.067$}    &\multirow{2}{*}{$0.070$} &\multirow{2}{*}{$0.070$}     &\multirow{2}{*}{$0.073$}    \\   
			
			\cline{2-5}
			&  P  & $462.5$ &$361.6$ & $4.157$&   &    &     & & &       & \\   \hline

			\multirow{4}{*}{Lung2}
			&\multirow{2}{*}{\parbox{0.2cm}{\centering{F+ C}}}  &\cellcolor{gray!20}  &  & \cellcolor{gray!20} &F     &  $4.437$    &$16.52$     &$0.702$ &$0.840$ &$14.54$        &$0.366$    \\   
			\cline{6-12} 
			
			& & \cellcolor{gray!20} \multirow{-2}{*}{$\textbf{0.008}$}   & \multirow{-2}{*}{$0.052$}     & \cellcolor{gray!20} \multirow{-2}{*}{$0.021$} &C     &  $0.012$    & $0.013$      &$0.013$ &$0.012$ &$0.010$        &$0.013$    \\   
			\cline{2-12}
			
			&T  &  $0.120$  &$782.4$    &$6.901$ &\multirow{2}{*}{T}     &  \multirow{2}{*}{$0.070$}    & \multirow{2}{*}{$0.067$}      &\multirow{2}{*}{$0.066$} &\multirow{2}{*}{$0.072$} &\multirow{2}{*}{$0.071$}        &\multirow{2}{*}{$0.070$}   \\   
			
			\cline{2-5}
			&  P  & $28.80$ &$0.905$ & $0.798$ &   &    &     & & &       & \\   \hline

			\multirow{4}{*}{Leukemia}
			
			&\multirow{2}{*}{\parbox{0.2cm}{\centering{F+ C}}} & \cellcolor{gray!20}  & \cellcolor{gray!20} & \cellcolor{gray!20} &F     &$1.793$    & $5.553$      &$0.649$ &$0.433$ &$1.079$        &$0.182$    \\   
			\cline{6-12} 
			
			& & \cellcolor{gray!20} \multirow{-2}{*}{$\textbf{0.001}$}   &  \cellcolor{gray!20} \multirow{-2}{*}{$\textbf{0.001}$}  & \cellcolor{gray!20} \multirow{-2}{*}{$0.002$} &C     &$0.337$    & $0.011$      &$0.013$ &$0.010$ &$0.010$        &$0.011$    \\  
			\cline{2-12}
			
			&T   & $0.103$  &   $8.043$    &$2.162$ &\multirow{2}{*}{T}      &\multirow{2}{*}{$0.270$}     &\multirow{2}{*}{$0.069$}      &\multirow{2}{*}{$0.068$}     &\multirow{2}{*}{$0.068$} &\multirow{2}{*}{$0.070$}     &\multirow{2}{*}{$0.072$}    \\   
			
			\cline{2-5}
			&  P  & $ 56.92$  & $1.000$ & $ 1.002$ &   &    &     & & &       & \\   \hline
			
			\multirow{4}{*}{Prostate}

			&\multirow{2}{*}{\parbox{0.2cm}{\centering{F+ C}}} &  \cellcolor{gray!20}  & \cellcolor{gray!20}  & \cellcolor{gray!20} &F     &$2.116$    & $6.010$      &$0.487$ &$0.362$ &$0.748$    & $0.167$    \\   
			\cline{6-12} 
			
			&   &\cellcolor{gray!20} \multirow{-2}{*}{$\textbf{0.001}$}  & \cellcolor{gray!20} \multirow{-2}{*}{$0.002$}  & \cellcolor{gray!20}  \multirow{-2}{*}{$0.002$} 
			&C     &$0.011$    & $0.012$      &$0.012$ &$0.010$ &$0.010$    & $0.010$    \\ 
			\cline{2-12}
			
			&T   & $0.183$ &  $7.088$    &$1.620$ 
			&\multirow{2}{*}{T}       &\multirow{2}{*}{$0.070$}   &\multirow{2}{*}{$0.067$}      &\multirow{2}{*}{$0.067$} &\multirow{2}{*}{$0.069$}     &\multirow{2}{*}{$0.071$}        &\multirow{2}{*}{$0.072$}    \\   
			
			\cline{2-5}
			&  P  &  $84.57$ & $0.822$ & $0.813$ &   &    &     & & &       & \\   \hline

			\multirow{4}{*}{Spambase}
			
			&\multirow{2}{*}{\parbox{0.2cm}{\centering{F+ C}}}  & \cellcolor{gray!20}   &   & &F     &$53.58$    & $65.03$      &$0.083$ &$38.79$ &$387.6$    &\cellcolor{gray!20} $0.049$    \\   
			\cline{6-12} 
			
			&  & \cellcolor{gray!20}   \multirow{-2}{*}{$\textbf{0.066}$}     & \multirow{-2}{*}{$0.237$}  & \multirow{-2}{*}{$0.482$}
			&C     &$0.016$    & $0.027$      &$0.031$ &$0.033$ &$0.033$    &\cellcolor{gray!20} $0.038$    \\ 
			\cline{2-12}
			
			&T   & $0.030$ &  $0.285$    &$0.012$ 
			&\multirow{2}{*}{T}                   &\multirow{2}{*}{$0.086$}    &\multirow{2}{*}{$0.006$}      &\multirow{2}{*}{$0.087$} &\multirow{2}{*}{$0.067$}      &\multirow{2}{*}{$0.065$}        &\multirow{2}{*}{$0.069$}    \\   
			
			\cline{2-5}
			&  P  & $2.321$  & $0.0283$ & $0.0296$  &   &    &     & & &       & \\   \hline

			\multirow{4}{*}{Dorothea}
			
			&\multirow{2}{*}{\parbox{0.2cm}{\centering{F+ C}}}  & \cellcolor{gray!20}  &  & \cellcolor{gray!20} &F     &$2152.7$    & $174610$      &$16.99$ &$53.52$ &$518.5$    &$217.1$    \\   
	    	\cline{6-12} 
			
			&    &\cellcolor{gray!20}  \multirow{-2}{*}{$\textbf{0.013}$}   & \multirow{-2}{*}{$0.035$}  & \cellcolor{gray!20} \multirow{-2}{*}{$0.033$}
			&C     &$0.051$    & $0.113$      &$0.102$ &$0.074$ &$0.027$    &$0.316$    \\ 
			\cline{2-12}
			
			&T    &$0.213$ &  $204.6$    &$7.208$ 
			&\multirow{2}{*}{T}      &\multirow{2}{*}{$0.005$}    &\multirow{2}{*}{$0.178$}      &\multirow{2}{*}{$0.069$} &\multirow{2}{*}{$0.067$}      &\multirow{2}{*}{$0.070$}        &\multirow{2}{*}{$0.067$}    \\   
			
			\cline{2-5}
			&  P  & $744.3$ & $15.06$ & $15.08$ &   &    &     & & &       & \\   \hline
			
			\multirow{4}{*}{News20}
			
			&\multirow{2}{*}{\parbox{0.2cm}{\centering{F+ C}}}  & \cellcolor{gray!20} & \cellcolor{gray!20}    & \cellcolor{gray!20} &F     &$--$        &$--$ &  $1444.8$   &$--$ &$--$        &$--$    \\   \cline{6-12} 
			
			& &\cellcolor{gray!20} \multirow{-2}{*}{$\textbf{9.190}$} &  \cellcolor{gray!20} \multirow{-2}{*}{$117.61$}  & \cellcolor{gray!20} \multirow{-2}{*}{$346.47$} &C     &$--$          &$--$ &  $36.26$ &$--$ &$--$        &$--$    \\  
			\cline{2-12}
			
			&T   & $0.5564$ &    $3076.1$    &$429.7$ &\multirow{2}{*}{T}    &\multirow{2}{*}{$--$}          &\multirow{2}{*}{$--$} &  \multirow{2}{*}{$0.106$} &\multirow{2}{*}{$--$} &\multirow{2}{*}{$--$}        &\multirow{2}{*}{$--$}    \\  
			
			\cline{2-5}
			&  P  & $2064.4$ & $1335.2 $ & $1331.0$&   &    &     & & &       & \\   \hline
			
			\textbf{Avg. rank} & & $1.23$ &$2.82$     &$3.23$    &     & $7.36$ &$8.73$ &$4.82$        &$5.27$  & $7.27$  &$4.27$      \\   \hline
		\end{tabular}
	\end{footnotesize}
	\vspace{-0.1in}
\end{table*}

\begin{table*}[!tb]
	\caption{Words (features) picked by IFC$^2$F are highlighted in yellow. The true/predicted label is given at the end of each review. The second column reports features selected for each review in ascending order (Y--axis) versus feature value (X--axis).}
	\label{tbl:interpret}
	\centering
	\footnotesize
	\begin{tabular}{|m{15.0cm}|m{2.3cm}|} \hline
		IMDB Review Text \textbf{(True Label, Predicted Label)} &  \\  \hline
		I work at a movie theater and every Thursday night we have an employee screening of one movie that comes out the next day. Today it was The Guardian. I saw the trailers and the ads and never expected much from it, and in no way really did i anticipate seeing this movie. Well turns out this movie was a lot more than I would have thought. It was a \hl{great} story first of all. Ashton Kutcher and Kevin Costner did amazing acting work in this film. Being a big fan of That 70's Show I always found it hard thinking of Kutcher as anyone but Kelso despite the \hl{great} acting he did in The Butterfly Effect, but after seeing this movie I think I might be able to finally look at him as a serious actor. It was also a \hl{great} tribute to the unsung heroes of the U.S. Coast Guard. \textbf{(positive, positive)}
		& \begin{minipage}{\textwidth}
			\includegraphics[width=2.4cm, height=1.7cm]{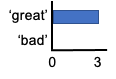}
		\end{minipage}
		\\ \hline		
		
		I saw this only because my 10-yr-old was bored. He and his friend hated it but of course liked being at the movies. This is the first time I've strongly disagreed with Ebert in many years. There is not a single thing to recommend this film. Willis is good, as always. But the story stinks, is unbelievable, there is \hl{no} real story, \hl{no} action, \hl{no} interesting cinematic sequences, \hl{no} surprises, and worst of all, the child star is A thoroughly repulsive slug guaranteed to turn off any parent who does not have a dweeby fat slob for a kid. By all means stay away and spare your child - unless you want to punish him or her. There is \hl{no} excuse for such lousy directing or writing and one hopes these filmmakers will suffer accordingly.
		 \textbf{(negative, negative) }
		& \begin{minipage}{\textwidth}
			\includegraphics[width=2.6cm, height=1.8cm]{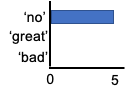}
		\end{minipage}
		\\ \hline		
		I felt compelled to comment on this film because it's listed as the fourth lowest-rated sci-film of all time on the IMDb. WHAT!?!? Sure, this movie is crappy, but it's HILARIOUS! It's not awful on an Ed Wood level, it's more surreal and uneven. There are some classic moments in the film. The brain surgery is gross and \hl{great} and \hl{even} nuttier when you consider that the film was rated PG! Gor chasing after his dolly before getting battery acid dumped on his face- "Mine! Gimmee!" Zandor Vorkoff's speeches at the beginning of the film- "Before Amir, Kali was but another weak nation struggling to break free from centuries of stagnant feudalism!" Angelo Rossito also has some \hl{great} lines- "\hl{No}, Gor! \hl{No}!" "You want these keys, don't you, my pretties?" It is absolutely wrong that this is the 4th lowest-rated sci-film on the IMDb because it is ENTERTAINING. \hl{No} matter how \hl{bad} a film is, if it still manages to be weird, quirky, unsettling, or entertaining, it has merit and doesn't deserve to be dumped on and dismissed. I won't defend most of Al Adamson's films, but this one, along with Dracula VS. FRANKENSTEIN and BLOOD OF GHASTLY HORROR, are entertaining enough to make up for their awfulness.
		\textbf{(positive, negative)}
		& \begin{minipage}{\textwidth}
			\includegraphics[width=2.6cm, height=2.7cm]{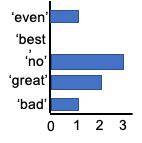}
		\end{minipage}
		\\ \hline		
		Jack Lemmon was one of our \hl{great} actors. His performances in Days Of Wine And Roses, The Apartment, Some Like It Hot, Missing (to name the first ones that come to mind) were all worthy of \hl{Best} Actor nomination. His only win was for Save The Tiger, and that's a shame. He gets melancholy down to a science, but never brings it into balance with the driver in his character. He actually did a similar character much better toward the end of his career in the one-note Glengarry Glen Ross. As for the movie, wonderful supporting work by Jack Gilford as Lemmon's partner and Thayer David as an arsonist, go for naught because the rest of the script is a muddled jumble of cliched vignettes, angst, neurotic nostalgia, and pointless moralizing. Worth seeing once as a time capsule into 1970's style experimental direction by Avildsen.
		 \textbf{(negative, positive)}
		& \begin{minipage}{\textwidth}
			\includegraphics[width=2.4cm, height=2.2cm]{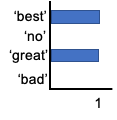}
		\end{minipage}
		\\ \hline		
	\end{tabular} 
\end{table*}

\subsection{Comparison with Baselines}\label{sec:baseline-comparison}

In this subsection, IFC$^2$F is compared with i) 2 dynamic feature selection methods: ETANA~\cite{liyanage2020fly}, F--ETANA~\cite{liyanage2020fly}, and ii) 6 streaming feature selection methods: OFS--Density~\cite{zhou2019ofs}, OFS--A3M~\cite{zhou2019online}, SAOLA~\cite{yu2014towards}, OSFS~\cite{wu2012online}, Fast--OSFS\cite{wu2012online}, and Alpha--Investing~\cite{zhou2005streaming}. 
In streaming feature selection methods~\cite{zhou2019ofs, zhou2019online, yu2014towards, wu2012online,zhou2005streaming}, a feature is selected if it satisfies an appropriatelly defined criterion (e.g., belongs in the approximated Markov blanket of the class variable ~\cite{yu2014towards, wu2012online}, $p$--statistic is greater than a dynamically varying threshold~\cite{zhou2005streaming}), or such that the boundary region of the decision is maintained as little as possible~\cite{zhou2019ofs, zhou2019online}. 
%
These methods are designed to handle sequentially arriving features during model training and select a \textit{global common} subset of features that is used to classify all instances during testing. Table \ref{tbl:complexity} summarizes the computational complexity of the baselines, as reported by their authors. The complexity of the proposed approach is discussed in Section \ref{sec:complexity_analysis}. The main reason for comparing with such methods is two--fold. First, both these methods and the proposed algorithm are sequential (i.e., examine one feature at a time). Second,  these baselines have been shown to outperform standard feature selection algorithms and scale well in high dimensional settings.
Similar to IFC$^2$F, ETANA and F--ETANA assume all features are available during training, while during testing, features arrive sequentially one at a time for each data instance. However, ETANA and F--ETANA assume features are conditionally independent given the class variable.

For a fair comparison, all streaming feature selection methods use a KNN classifier with three neighbors to evaluate a selected feature subset, since it has been shown to outperform SVM, CART, and J48 classifiers on the datasets used in~\cite{zhou2019ofs, yu2014towards}. At the same time, parameter $\alpha$ used by SAOLA, OSFS, and Fast--OSFS is set to  $0.01$, which has been shown to produce the best performance \cite{yu2014towards, wu2012online}. The code for all baselines is either publicly available or has been provided by their authors. The same training and testing datasets are used by all methods. Finally, the same metrics (i.e., accuracy, number of features used, time) used by the baselines are adopted.
Observations from Tables~\ref{tbl:acc},~\ref{tbl:feat},~\ref{tbl:time} are summarized next. \\
\noindent\textbf{Madelon:} IFC$^2$F achieves the highest accuracy. In fact, this corresponds to an improvement of $7.8\%$ in accuracy with being $27.8\%$ faster in joint feature selection and classification compared to ETANA, which has the second highest accuracy.  ETANA, however, requires $18.7\%$ less features on average compared to IFC$^2$F. \\
\noindent \textbf{Lung Cancer:} ETANA, F--ETANA, SAOLA, and Fast--OSFS achieve the highest accuracy, but require $50.4\%$ to $ 3.75$ $\times$ $10^{3}\%$ more features on average and are $200\%$ to $1.82$ $\times$ $10^{6}\%$  slower in joint feature selection and classification for a difference of $1.7\%$ in accuracy  compared to  IFC$^2$F. \\
\noindent \textbf{MLL:} Both IFC$^2$F and ETANA achieve $100\%$ accuracy. However, IFC$^2$F requires $32.9\%$ less features on average and is $200\%$ faster in joint feature selection and classification compared to ETANA. \\
\noindent \textbf{Dexter:}  IFC$^2$F  achieves the highest accuracy and is the fastest in joint feature selection and classification. \\
\noindent \textbf{Car:} F--ETANA achieves the highest accuracy ($10.7\%$ improvement), but requires $1.31$ $\times$ $10^{3}\% $ more features on average and is $1.26$ $\times$ $10^{3}\% $ slower in joint feature selection and classification compared to  IFC$^2$F. \\
\noindent \textbf{Lung2:}  OFS--Density achieves the highest accuracy, but requires $43.4\%$ more features on average and is $5.55$ $\times$ $10^{4}\%$ slower in joint feature selection and classification compared to IFC$^2$F for a difference of  $5.4\%$ in accuracy. \\
\noindent \textbf{Leukemia:} IFC$^2$F, ETANA and F--ETANA achieve the highest accuracy. However, IFC$^2$F requires $8.7\%$ and $80.1\%$ less features on average compared to ETANA and F--ETANA, respectively.  \\
\noindent \textbf{Prostate:} ETANA achieves the highest accuracy, but is $50\%$ slower in joint feature selection and classification compared to IFC$^2$F. \\
\noindent \textbf{Spambase:} OFS--A3M achieves the highest accuracy, but requires $\sim$ $9$ times more features for a difference of $4.9\%$ in accuracy and is much slower compared to IFC$^2$F. \\
\noindent \textbf{Dorothea:} IFC$^2$F and Fast--OSFS achieve the highest accuracy. However, Fast--OSFS requires $\sim$ $10$ times more features on average and is much slower compared to  IFC$^2$F.

For the majority of datasets, ETANA achieves the highest accuracy while IFC$^2$F is competitive achieving higher or closely second performance compared to ETANA. This result demonstrates the fact that for some datasets, the assumption of features being conditionally independent given the class variable (used in ETANA) is more appropriate than assuming that features are dependent (used in IFC$^2$F). On the other hand,  IFC$^2$F requires less number of features on average than ETANA, while OSFS consistently selects the least number of features among all baselines. This observation  suggests that considering feature dependencies helps to get rid of redundant features.  Further, IFC$^2$F is the fastest algorithm to perform joint feature selection and classification in all the datasets compared to the baselines. This is due to the fact that it uses less number of features on average per data instance. Specifically, easy to classify data instances require few features as opposed to more challenging data instances that require more features
to be accurately classified by IFC$^2$F.

\subsection{Performance Assessment on a High Dimensional Dataset}\label{sec:comparison-high-dimension-dataset}

In this subsection, the performance of IFC$^2$F and the baselines is discussed within the context of the News20 dataset. Except for IFC$^2$F, ETANA, F--ETANA and SAOLA, the rest of the methods were unable to generate results within a cutoff time of 12 days. Although SAOLA achieves the highest accuracy, it requires $\sim$6 times more features and is $\sim$160 times slower in joint feature selection and classification for a mere improvement of $4.6\%$ in accuracy compared to IFC$^2$F (second last row in Tables~\ref{tbl:acc},~\ref{tbl:feat},~\ref{tbl:time}).  This experiment demonstrates the ability of IFC$^2$F  to scale for more than 1.3 million features.

\subsection{Statistical Significance}

To validate the statistical significance of the results presented in Sections \ref{sec:baseline-comparison} and \ref{sec:comparison-high-dimension-dataset}, a Friedman test, which constitutes a well--known method to compare the performance of several algorithms across multiple datasets~\cite{demvsar2006statistical}, is conducted. The average ranking (avg. rank) of each method is given  in the last row in Tables \ref{tbl:acc}--\ref{tbl:time}. The $p$--values of the Friedman test  on classification accuracy, the average number of features used and time required for joint feature selection and classification are $1.69\times10^{-4}$, $8.07\times10^{-7}$ and $4.55\times10^{-24}$, respectively. Thus, there is a significant difference~\cite{demvsar2006statistical} in the performance of IFC$^2$F and the baselines.

\begin{figure}[!tb]
	\centering
	\includegraphics[width=0.98\columnwidth]{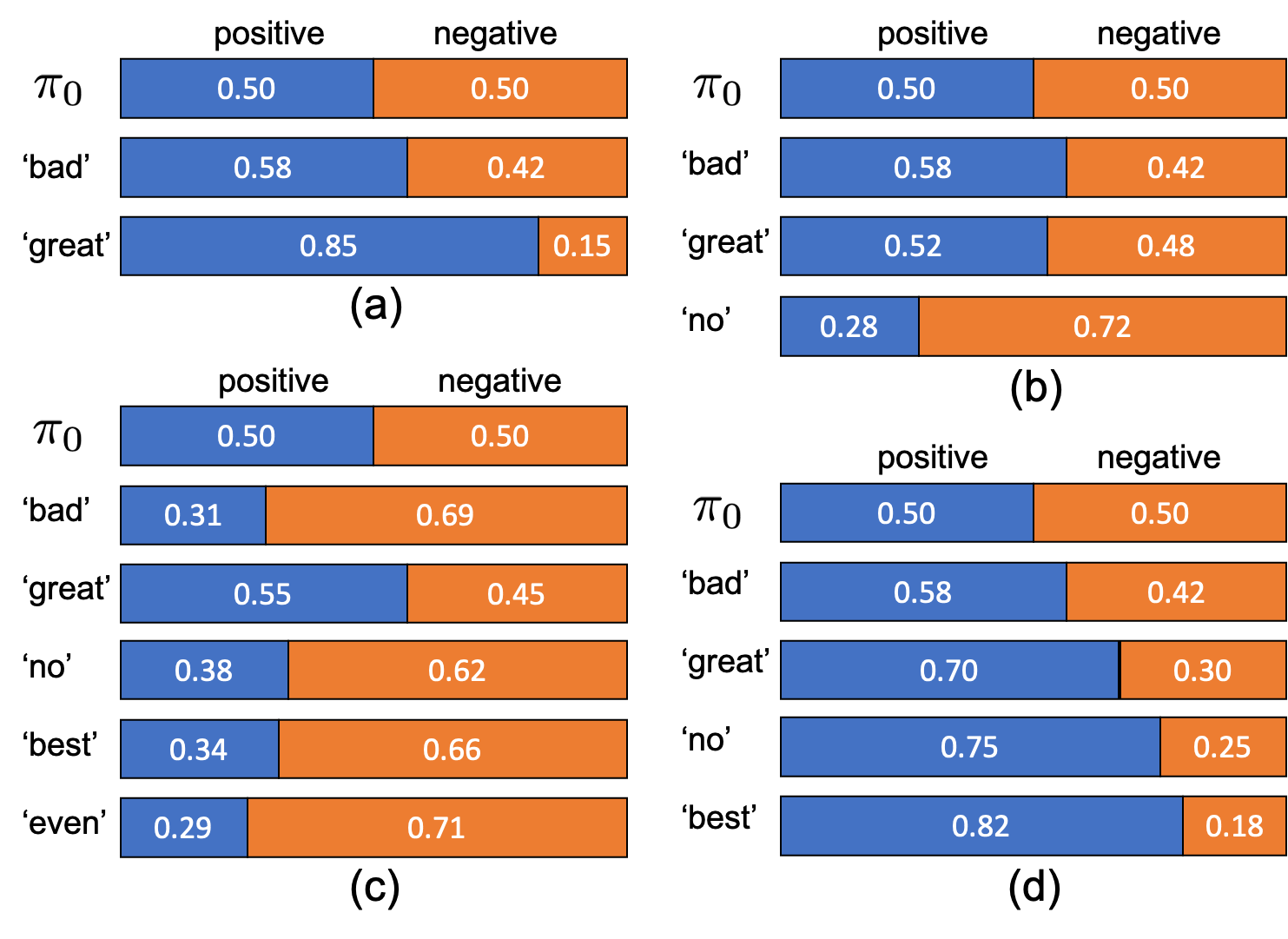} 
	\caption{Evolution of posterior probability distribution, i.e., $P(\text{positive}|F_1,\dots, F_k)$ (in blue) and $P(\text{negative}|F_1,\dots, F_k)$ (in orange) for 4 IMDB reviews in Table~\ref{tbl:interpret}.}
	\label{fig:instances}
\end{figure}

\subsection{Demonstration of Instance--wise Feature Selection}\label{sec:instance-wise_feature_selection}

Table~\ref{tbl:interpret} demontrates the instance--wise nature of IFC$^2$F using $4$ illusrative examples from the IMDB movie reviews dataset ($50,000$ instances, $89,523$ features, $2$ classes)~\cite{maas2011learning}. We have selected the IMDB dataset, because the raw text of reviews is available and can be directly used to interpret the classification rationale. The training and validation sets with bag--of--words features are used as provided. The Markov blanket based feature ordering is $\{$ `bad', `great', `no', `best', `even', `plot', `nothing', `love', `don't', `waste', $\dots \}$. Figure~\ref{fig:instances} illustrates the evolution of the posterior probability distribution $\pi_k$ as more and more features are evaluated, until the stopping condition $g(\pi_k) \leqslant \mathcal{A}_k (\pi_k)$ is satisfied. At that time, the instance is assigned to the class with the maximum posterior probability; this is a direct result of using constant missclassification costs, i.e., $Q_{01}=Q_{10} =1, Q_{00} =Q_{11} = 0$ (see Eq.~(\ref{eq:optimal_classification_strategy})). Observe that the proposed framework evaluates more features to predict challenging reviews such as (c) and (d) compared to easy and straightforward reviews such as (a) and (b).  In summary, IFC$^2$F selects \textit{different features} for \textit{different data instances} in a dynamic setting and assigns the class label based on the observed features.


\begin{figure}[t]
	\centering
	\includegraphics[width=\columnwidth]{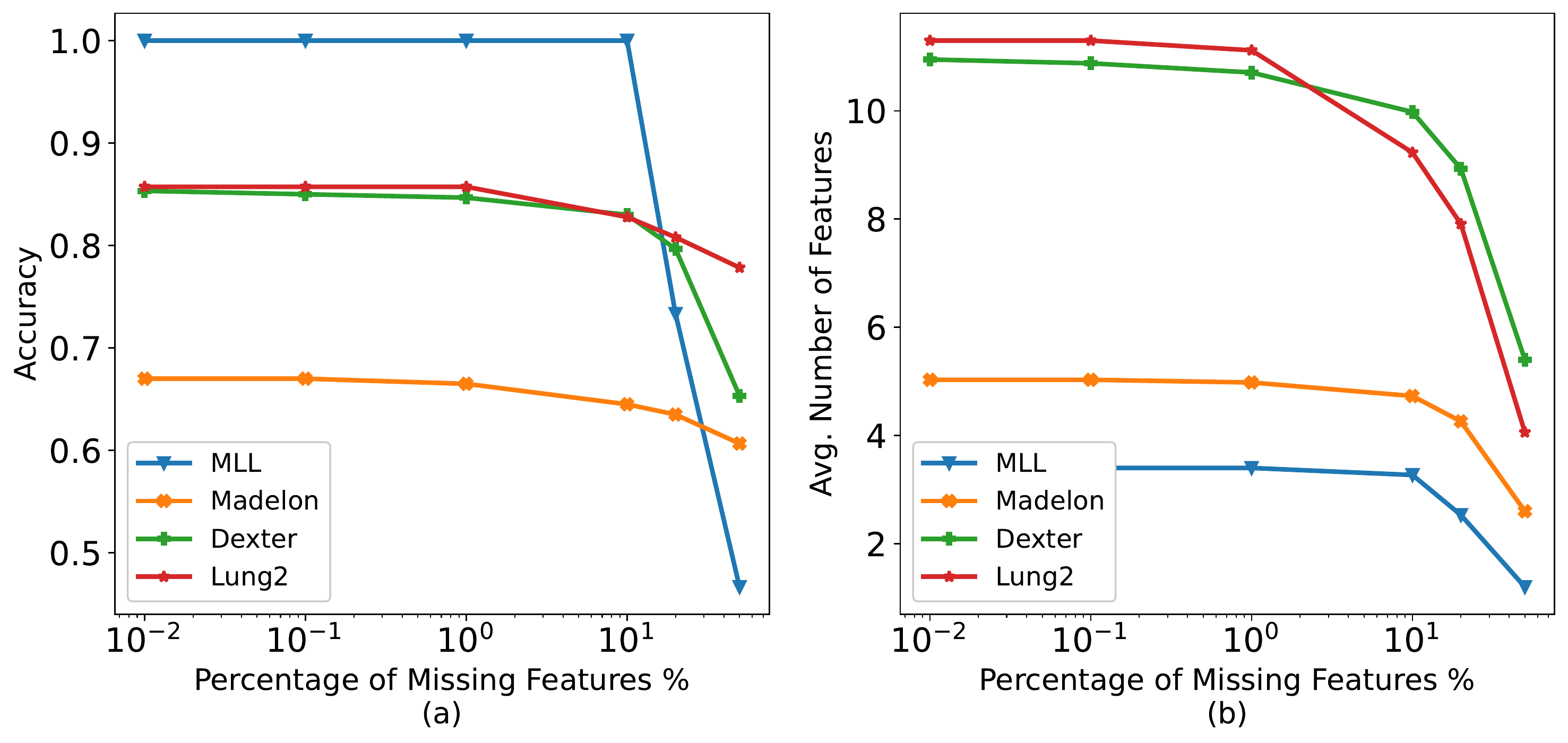} 
	\vspace{-0.1in}
	\caption{Variation of (a) accuracy, and (b) average number of features as the percentage of missing features increases from $0.01\%$ to $50\%$ across datasets.}
	\label{fig:missing_feat}
	\vspace{-0.1in}
\end{figure}

\subsection{Robustness to Missing Features}

In this subsection, the ability of IFC$^2$F to handle missing features is demonstrated.
Specifically, $x\%$ of features are randomly removed from each test instance and the posterior probability $\pi_k$ is kept unchanged if a feature is missing. $x\%$ is increased from $0.01\%$ to $50\%$ and the effect on the accuracy and the average number of features used for classification is noted (see Fig~\ref{fig:missing_feat}). Note that  the accuracy achieved by IFC$^2$F is robust for up to $10\%$ missing features. Thus, IFC$^2$F can identify informative features to make accurate predictions even if some important features may be missing. 

\section{Conclusion and Future Directions}

In this paper, a framework to perform dynamic instance--wise joint feature selection and classification with correlated features is proposed. Specifically, feature dependencies are modeled using a Bayesian network. Based on the learned dependency network, a method is proposed to sequentially select the most informative features and reach a classification decision for each instance individually. The effectiveness and scalability of the proposed method is illustrated on various real--world datasets. The proposed method robustly performs well on all of them, with comparable and often superior performance compared to prior art.

The proposed method selects the most informative features from the dependency graph utilizing the proposed Markov blanket based feature ordering. This dependency graph, however, is learned offline during training, hence the ordering in which features are selected is common for all test instances. In future work, the goal is to address this limitation by dynamically learning the network structure, since the number of selected features and the interpretability of the classification depends on the graph structure at hand. At the same time, to keep the preprocessing time small, the proposed method opts for filtering out features based on the mutual information between each feature and the class label. In future work, approaches such as multivariate mutual information can be explored to better capture feature dependencies. Lastly, the proposed method assumes all data instances are available at once during training, which may not hold in cases where data instances are provided sequentially. In the future, the applicability of online learning methods \cite{hoi2018online} in the proposed setting can be considered.

\section*{Appendix}
\subsection{Proof of Theorem~\ref{thm:dp_programming}}
At the end of the $K$th stage, assuming that all the features have been examined, the only remaining expected cost is the optimum misclassification cost of selecting among $L$ classes, which is $\bar{J}_{K} (\pi_{k} ) = g(\pi_{k} )$.

Then, consider any intermediate stage $k=0,1,\dots,K-1$. Being at stage $k$, with available information $\pi_{k}$, the optimum strategy has to choose between, either to terminate and incur cost $g (\pi_{k} )$, which is the optimum misclassification cost of selecting among $L$ classes, or continue with the next feature $F_{k+1}$, and incur cost $e_{k+1}$ and an additional cost $\bar{J}_{k+1}  (\pi_{k+1} )$ to continue optimally at stage $k+1$. Thus, the total cost of continuing optimally is $e_{k+1}+\bar{J}_{k+1} ( \pi_{k+1} )$.
However, at stage $k$, the assignment  $f_{k+1}$ of the next feature $F_{k+1}$ is not known. 
Thus, the expected \textit{optimum cost--to--go}, which is equal to $ e_{k+1}+  \mathbb{E} \big \lbrace\bar{J}_{k+1} ( \pi_{k+1} )|\pi_{k} \big \rbrace$, needs to be considered. Using Bayes' rule to express $\pi_{k+1}$ in terms of $\pi_{k}$, and by the definition of the expectation operator, the \textit{optimum cost--to--go} $\mathcal{A}_{k} ( \pi_{k})$ takes the following form:
\begin{align}
\mathcal{A}_{k} (\pi_{k} ) &\triangleq e_{k+1}+  \mathbb{E} \big \lbrace\bar{J}_{k+1} ( \pi_{k+1} )|\pi_{k} \big \rbrace \nonumber \\
\label{eq.dynamic0}
&= e_{k+1} + \sum_{F_{k+1}} P(F_{k+1}|F_{1},\dots,F_{k}) \nonumber \\
&\times \bar{J}_{k+1} \bigg (  \frac{\diag \big(\Delta(F_{k+1}|F_{1},\dots,F_{k},\mathcal{C}) \big) \pi_{k} }{\Delta^T(F_{k+1}|F_{1},\dots,F_{k},\mathcal{C}) \pi_{k}} \bigg ). 
\end{align}
\noindent Next, the term $P \big(F_{k+1}|  F_{1},F_{2},\dots,F_{k} \big)$ must be simplified. Specifically, using the Bayes' rule and the law of total probability, it can be shown that:
\begin{align}
P \big(F_{k+1}|F_{1},F_{2},\dots,F_{k} \big)  &= \frac{P(F_{1},F_{2},\dots,F_{k+1})}{P(F_{1},F_{2},\dots,F_{k})} \nonumber \\
&= \frac{\sum_{j = 1}^L P(F_{1}, \dotsc, F_{k+1} \mathcal{C}_j)}{\sum_{j = 1}^L P(F_{1}, \dotsc, F_{k},\mathcal{C}_j)} \nonumber \\
\label{eq.dynamic3}
&= \frac{\sum_{j = 1}^L P(F_{1}, \dotsc, F_{k+1}| \mathcal{C}_j)P(\mathcal{C}_j) }{\sum_{j = 1}^L P(F_{1}, \dotsc, F_{k} | \mathcal{C}_j)P(\mathcal{C}_j)}.	
\end{align}
\noindent Using the chain rule, Eq.~(\ref{eq.dynamic3}) can  be simplified as follows:
\begin{align}
&P \big( F_{k+1}|F_{1},F_{2},\dots,F_{k} \big) \nonumber \\  
&= \frac{ \sum_{j = 1}^L P(F_{1}, \dotsc, F_{k} | \mathcal{C}_j)
	P( F_{k+1} | F_{1}, \dotsc, F_{k},\mathcal{C}_j) P(\mathcal{C}_j) }
{\sum_{j = 1}^L P(F_{1}, \dotsc, F_{k} | \mathcal{C}_j)P(\mathcal{C}_j)} \nonumber \\
&= \sum_{j = 1}^L  \frac{  P(F_{1}, \dotsc, F_{k} | \mathcal{C}_j)
	P(\mathcal{C}_j) }
{\sum_{j = 1}^L P(F_{1}, \dotsc, F_{k} | \mathcal{C}_j)P(\mathcal{C}_j)} P( F_{k+1} | F_{1}, \dotsc, F_{k},\mathcal{C}_j)   \nonumber \\
&= \sum_{j = 1}^L  P(\mathcal{C}_j| F_{1}, \dotsc, F_{k} ) P( F_{k+1} | F_{1}, \dotsc, F_{k},\mathcal{C}_j)   \nonumber \\
&= \sum_{j = 1}^L \pi_{k}^j P( F_{k+1} | F_{1}, \dotsc, F_{k},\mathcal{C}_j) \nonumber \\
\label{eq.dynamic2}
&=  \Delta^T(F_{k+1}|F_{1},\dots,F_{k},\mathcal{C}) \pi_{k}.
\end{align}
\noindent Finally, substituting Eq.~(\ref{eq.dynamic2}) in Eq.~(\ref{eq.dynamic0}), the desired result can be acquired:
\begin{align}
\mathcal{A}_{k} ( \pi_{k+1} ) &=  e_{k+1} + \sum_{F_{k+1}}  \Delta^T(F_{k+1}|F_{1},\dots,F_{k},\mathcal{C})   \pi_{k} \nonumber \\
&  \times \bar{J}_{k+1} \bigg (  \frac{\diag(\Delta(F_{k+1}|F_{1},\dots,F_{k},\mathcal{C})) \pi_{k}}{\Delta^T(F_{k+1}|F_{1},\dots,F_{k},\mathcal{C}) \pi_{k}} \bigg ),
\end{align}
which completes the proof.

\subsection{Proof of Lemma~\ref{lem:g_function}}

Consider the definition of $g(\varpi)$:
\begin{equation*}
g(\varpi)\triangleq \min_{1 \leqslant j \leqslant L} \big [   Q_j^T \varpi \big ], \varpi \in [0,1]^{L}.
\end{equation*}
The term $ Q_j^T \varpi$ is linear with respect to $\varpi$, and since the minimum of linear functions is a concave, piecewise linear function, $g(\varpi)$ is
a concave, piecewise linear function as well. Concavity also assures the continuity of this function. Minimization over finite $L$ hyperplanes guarantees that the function $g(\varpi)$ is made up of at most $L$ hyperplanes. Hence the set $\{ Q_j^T \}_{j=1}^L$ of $L$ vectors represents those $L$ hyperplanes.

\subsection{Proof of Lemma~\ref{lem:A_function}}

First, consider the function $\mathcal{A}_{K-1} (\varpi)$ given by: 
\begin{align}
\label{eq:convexity1}
\mathcal{A}_{K-1} (\varpi) &= e_K + \sum_{F_{K}}  \Delta^T(F_{K}|F_{1}, \dots, F_{K-1},\mathcal{C}) \varpi \nonumber \\
&\times \bar{J}_{K} \bigg (  \frac{\diag \big(\Delta (F_{K}|F_{1},\dots,F_{K-1},\mathcal{C} ) \big) \varpi }{\Delta^T(F_{K}|F_{1},\dots,F_{K-1},\mathcal{C})\varpi }\bigg ). 
\end{align}
Using the fact that $\bar{J}_{K}(\pi_{K}) = g(\pi_{K})$, Eq.~(\ref{eq:convexity1}) can be rewritten as follows: 
\begin{align}
\label{eq:convexity2}
\mathcal{A}_{K-1} (\varpi) &= e_K + \sum_{F_{K}}  \Delta^T(F_{K}|F_{1}, \dots, F_{K-1},\mathcal{C}) \varpi  \nonumber \\ &\times g \bigg (  \frac{\diag \big(\Delta (F_{K}|F_{1},\dots,F_{K-1},\mathcal{C} ) \big) \varpi }{\Delta^T(F_{K}|F_{1},\dots,F_{K-1},\mathcal{C})\varpi }\bigg ).  
\end{align}
Using the definition of $g(\varpi)$, Eq.~(\ref{eq:convexity2}) can be rewritten as follows:
\begin{align}
\mathcal{A}_{K-1} (\varpi) &= e_K + \sum_{F_{K}}  \Delta^T(F_{K}|F_{1}, \dots, F_{K-1},\mathcal{C}) \varpi  \nonumber \\  
&\label{eq:convexity7}  \times \min_{1 \leqslant j \leqslant L} \Bigg[  \frac{Q_j^T\diag \big(\Delta (F_{K}|F_{1},\dots,F_{K-1},\mathcal{C} ) \big) \varpi }{\Delta^T(F_{K}|F_{1},\dots,F_{K-1},\mathcal{C})\varpi } \Bigg ].   
\end{align}
Using the facts that $Q_j$ and  $\Delta \big(F_{K}|F_{1},\dots,F_{K-1},\mathcal{C} \big) $ are  non--negative  vectors, Eq.~(\ref{eq:convexity7}) can be simplified as follows:
\begin{align}
\mathcal{A}_{K-1} (\varpi) &= e_K\nonumber \\ \label{eq:convexity4}
&+ \sum_{F_{K}} \min_{1 \leqslant j \leqslant L} \Big[ Q_j^T\diag \big(\Delta (F_{K}|F_{1},\dots,F_{K-1},\mathcal{C} ) \big) \varpi \Big ].
\end{align}
Note that the term $ Q_j^T\diag \big(\Delta (F_{K}|F_{1},\dots,F_{K-1},\mathcal{C} ) \big) \varpi $ is linear with respect to $\varpi$. Using the facts that i) $e_K >0$, ii) the minimum of linear functions is a concave, piecewise linear function, and iii) the non--negative sum of concave/piecewise linear functions is also a concave/piecewise linear function, implies that $\mathcal{A}_{K-1}(\varpi) $ is a concave, piecewise linear function.  Concavity also assures the continuity of this function.

Then, consider the function $\mathcal{A}_{K-2}(\varpi)$ given by: 
\begin{small}
	\begin{align}
	\label{eq:convexity5}
	\mathcal{A}_{K-2}(\varpi) &= e_{K-1}+ \sum_{F_{K-1}}  \Delta^T(F_{K-1}|F_{1},\dots, F_{K-2},\mathcal{C}) \varpi  \nonumber \\ 
	&\times \bar{J}_{K-1} \bigg (  \frac{\diag \big(\Delta(F_{K-1}|F_{1},\dots,F_{K-2},\mathcal{C}) \big) \varpi }{\Delta^T(F_{K-1}|F_{1},\dots,F_{K-2},\mathcal{C})
		\varpi }\bigg ). 
	\end{align}
\end{small}\normalsize

Note that $\bar{J}_{K-1} (\varpi) = \min \big[g(\varpi), \mathcal{A}_{K-1} (\varpi) \big]$ (see Theorem~\ref{thm:dp_programming}). Using the facts that i) $g(\varpi)$ is a concave, piecewise linear function, ii) $\mathcal{A}_{K-1} (\varpi)$ is a concave, piecewise linear function, and iii) the minimum of two concave/piecewise linear functions is also a concave/piecewise linear function, implies that $\bar{J}_{K-1} (\varpi)$ is also concave and piecewise linear. Furthermore, the non--negative sum of concave/piecewise linear functions is also a concave/piecewise linear function. Based on this fact and the facts that $e_{K-1}>0$, and $\Delta(F_{K-1}|F_{1},\dots,F_{K-2},\mathcal{C})$ is a non--negative vector, the function  $\mathcal{A}_{K-2}(\varpi)$ is concave and piecewise linear. Concavity also assures the continuity of this function. Using similar arguments, the concavity, the continuity and the piecewise linearity of functions $\mathcal{A}_{k}(\varpi), k=0,\dots,K-3$, can also be guaranteed.

\subsection{Proof of Theorem~\ref{thm:dynamic_proper}}

At the final stage, i.e., $k = K$, $\bar{J}_{K}(\varpi) = g(\varpi) = \min_{1 \leqslant j \leqslant L} \big [   Q_j^T \varpi \big ].$
Hence, $\{\alpha_K^i\} \triangleq  \{Q_j^T\}_{j=0}^L$. The rest of the proof is very intuitive.  Using the facts that i) $g(\varpi)$ and $ \mathcal{A}_{k}(\varpi)$ are  concave and piecewise linear with respect to $\varpi$, ii) $\bar{J}_{k}(\varpi) = \min \big[ g(\varpi), \mathcal{A}_{k}(\varpi) \big], k\in\{0,\dots,K-1\} $  (see Theorem~\ref{thm:dp_programming}), and iii) the minimum of two concave/piecewise linear functions is also a concave/piecewise linear function, implies that the function $\bar{J}_{k}(\varpi)$ is also concave and piecewise linear. Finally, since $\bar{J}_{k}(\varpi)$ is a concave and piecewise linear function defined on a probability space, it is noted that $\bar{J}_{k}(\varpi) = \min_{i}  [\alpha_k^i\varpi]$, where the set $\{\alpha_k^i\}_i$ of vectors represents its linear pieces. 

\bibliographystyle{IEEEtran}
\begin{small}
	\bibliography{IEEEabrv,references}
\end{small}

\begin{IEEEbiography}[{\includegraphics[width=1.16in,height=1.16in,clip,keepaspectratio]{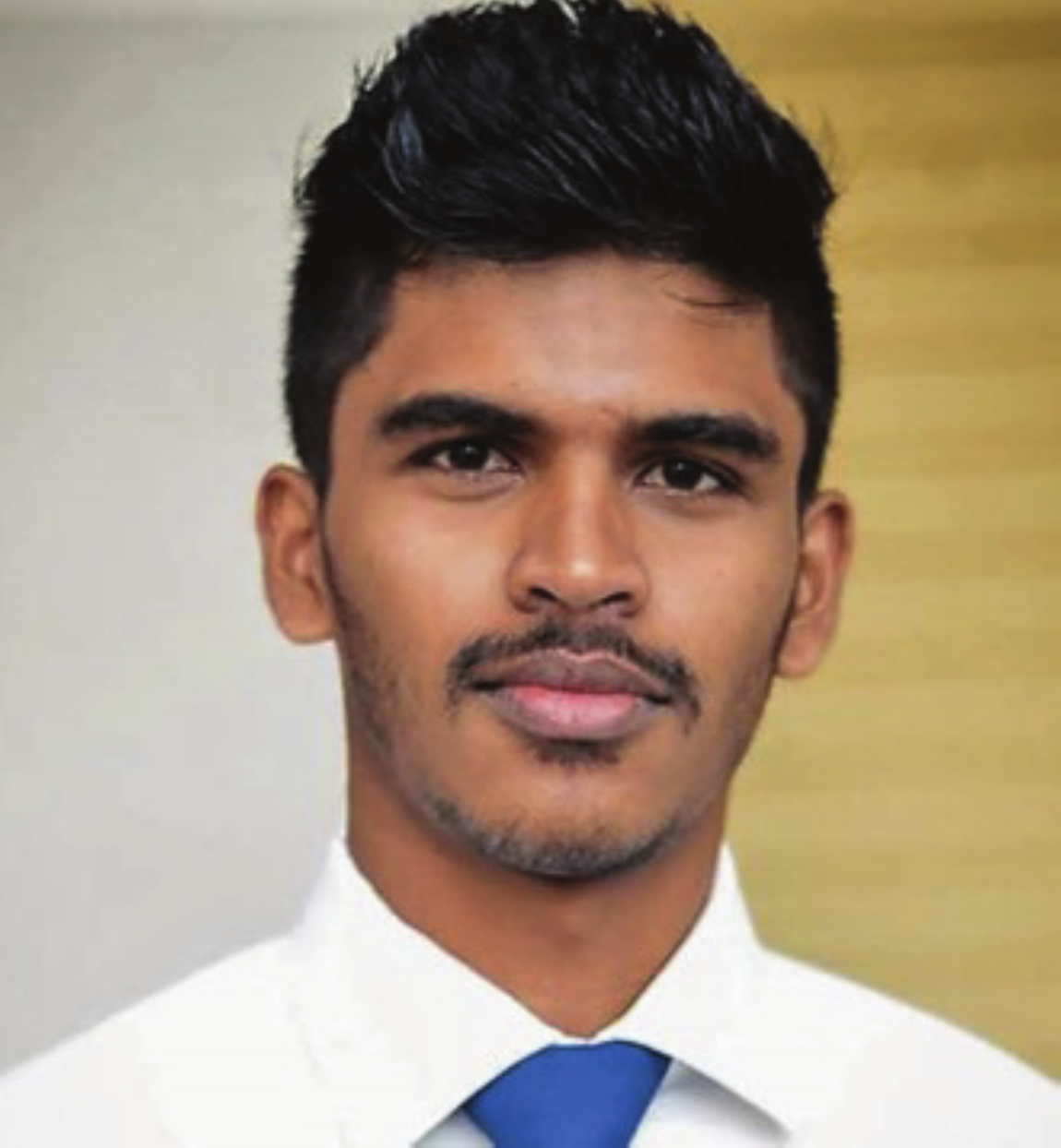}}]
	{Yasitha Warahena Liyanage} received the B.S. degree in electrical and electronic engineering from the University of Peradeniya, Sri Lanka, in 2016. Currently, he is working toward the Ph.D. degree in electrical and computer engineering at the University at Albany, SUNY. His research interests include quickest change detection, optimal stopping theory and machine learning.
\end{IEEEbiography}
\begin{IEEEbiography}[{\includegraphics[width=1.25in,height=1.25in,clip,keepaspectratio]{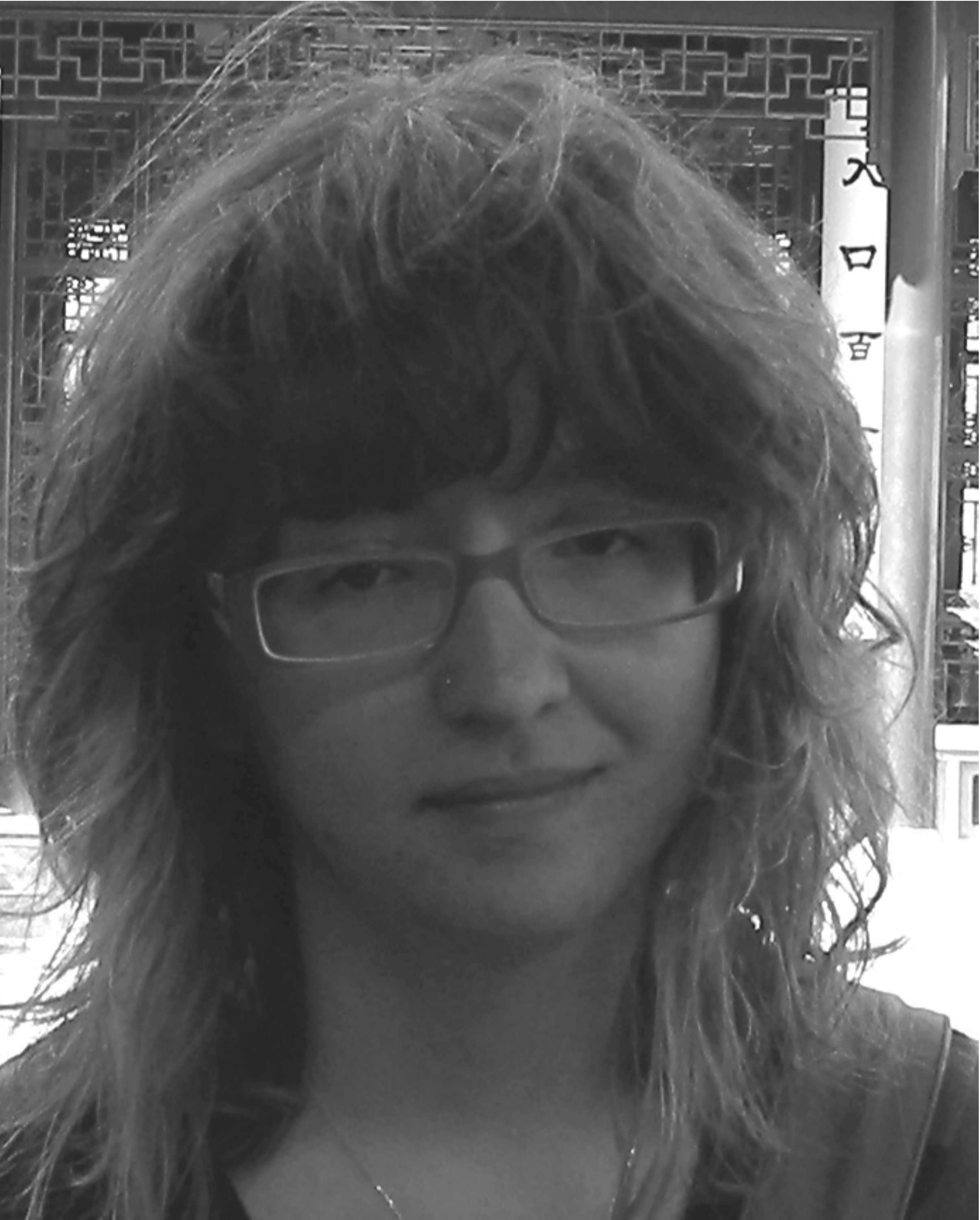}}]
	{Daphney-Stavroula Zois} received the B.S. degree in computer engineering and informatics from the University of Patras, Patras, Greece, and the M.S. and Ph.D. degrees in electrical engineering from the University of Southern California, Los Angeles, CA, USA. Previous appointments include the University of Illinois, Urbana--Champaign, IL, USA. She is an Assistant Professor in the Department of Electrical and Computer Engineering, University at Albany, State University of New York, Albany, NY, USA. She received the Viterbi Dean's and Myronis Graduate Fellowships, and the NSF CAREER award. She has served and is serving as Co--Chair, TPC member or reviewer in international conferences and journals, such as AAAI, ICASSP, GlobalSIP, Globecom, and IEEE Transactions on Signal Processing. Her research interests include decision making under uncertainty, machine learning, detection \& estimation theory, intelligent systems design, and signal processing.
\end{IEEEbiography}
\begin{IEEEbiography}[{\includegraphics[width=1.2in,height=1.2in,clip,keepaspectratio]{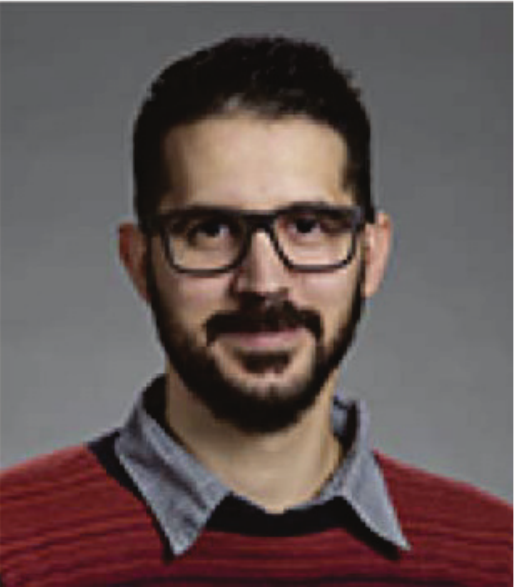}}]
	{Charalampos Chelmis} is an Assistant Professor in Computer Science at the University at Albany, State University of New York, and the director of the Intelligent Big Data Analytics, Applications, and Systems (IDIAS) Lab, focusing on problems involving big, often networked, data. He has served and is serving as Co-Chair, TPC member or reviewer in numerous international conferences and journals such as ASONAM, SocInfo, and ICWSM. Currently, he serves as Associate Editor of the Journal of Parallel and Distributed Systems, and served as Guest Editor for the Encyclopedia of Social Network Analysis and Mining. He received the B.S. degree in computer engineering and informatics from the University of Patras, Greece in 2007, and the M.S. and Ph.D. degrees in computer science from the University of Southern California in 2010 and 2013, respectively.
\end{IEEEbiography}
	
	

\end{document}